\documentclass[10pt,twocolumn,letterpaper]{article}

\usepackage{cvpr}              % To produce the CAMERA-READY version
\definecolor{cvprbluelinkcolor}{rgb}{0.21,0.49,0.74}
\definecolor{darkblue}{RGB}{0, 0, 200}
\definecolor{purple}{rgb}{0.51, 0.255, 0.51}
\definecolor{red}{rgb}{100, 0.0, 0}
\definecolor{green}{RGB}{0, 153, 76}
\definecolor{mag}{RGB}{204,0, 204}
\usepackage[pagebackref,breaklinks,colorlinks,allcolors=cvprbluelinkcolor=mag,
anchorcolor=darkblue,
urlcolor=darkblue,
citecolor=darkblue]{hyperref}
\usepackage{url}
\usepackage{subcaption}
\usepackage{enumitem}
\usepackage{tcolorbox}
\usepackage{booktabs,multirow,colortbl,xcolor}
\definecolor{BrickGreen}{RGB}{0,145,0}
\definecolor{BrickRed}{RGB}{180,0,0}
\newcommand{\inc}[1]{{\color{BrickGreen}(#1)}} 
 
\newcommand{\incr}[1]{{\color{BrickGreen}#1}} 
\newcommand{\decr}[1]{{\color{BrickRed}#1}} 
\usepackage{graphicx}
\usepackage{cuted}
\usepackage[capitalize]{cleveref}
\usepackage{kotex}
\usepackage{wrapfig}
\usepackage{siunitx}  
\usepackage{amsmath, amssymb}
\usepackage{amsthm}
\usepackage{algorithm}
\usepackage{algpseudocode}

\newcommand{\appTocSec}[1]{%
  \par\addvspace{0.45em}%
  \noindent
  \makebox[1.8em][l]{\textbf{\ref*{#1}}}%
  \textbf{\nameref*{#1}}%
  \nobreak\leaders\hbox{\kern0.35em.\kern0.35em}\hfill%
  \nobreak\hyperref[#1]{\textbf{\pageref*{#1}}}%
  \par
}
\newcommand{\appTocSubManual}[3]{%
  \par\addvspace{0.15em}%
  \noindent\hspace{2.3em}%
  \makebox[3.8em][l]{#1}%
  #3%
  \nobreak\leaders\hbox{\kern0.35em.\kern0.35em}\hfill%
  \nobreak\hyperref[#2]{\pageref*{#2}}%
  \par
}

\newtheorem{lemma}{Lemma}
\def\paperID{12427} % *** Enter the Paper ID here
\def\confName{CVPR}
\def\confYear{2026}

\title{Activation Quantization of Vision Encoders Needs Prefixing Registers}

\author{
Seunghyeon Kim$^1$,
Taesun Yeom$^1$, 
Jinho Kim$^2$,
Wonpyo Park$^{3}$,
Kyuyeun Kim$^{3}$,
Jaeho Lee$^{1}$\thanks{Corresponding Author.} \\
POSTECH$^1$, NYU Langone Health$^2$, Google$^3$\\
{\tt\small $\{$shkim0418, tsyeom, jaeho.lee$\}$@postech.ac.kr, Jinho.Kim@nyulangone.org}\\
{\tt\small $\{$wppark, kyuyeunk$\}$@google.com}
}

\begin{document}
\maketitle
\begin{abstract}
Large pretrained vision encoders are central to multimodal intelligence, powering applications from on-device vision processing to vision-language models. Since these applications often demand real-time processing of massive visual data, reducing the inference cost of vision encoders is critical. Quantization offers a practical path, but it remains challenging even at 8-bit precision due to so-called outliers. In this work, we propose \textit{RegCache}, a training-free algorithm that mitigates outliers in large-scale pretrained vision encoders and serves as a plug-in module that can be applied on top of other quantization methods. RegCache introduces outlier-prone yet semantically meaningless prefix tokens into the vision encoder, which prevent other tokens from having outliers. Notably, we observe that outliers in vision encoders behave differently from those in language models, motivating two technical innovations: middle-layer prefixing and token deletion. Experimental results show that our method consistently improves quantized model performance across various vision encoders, particularly in extremely low-bit regimes (e.g., 4-bit). 

\end{abstract}    
\section{Introduction}\label{sec:introduction} 
Transformer-based vision encoders, such as CLIP and DINOv2, lie at the core of modern multimodal systems \cite{radford2021learning,oquab2024dinov}. Leveraging the scalability of vision transformers (ViTs) \cite{dosovitskiy2021an}, these models are pretrained on massive datasets using large-scale computation, yielding highly informative and versatile visual representations. However, their large size poses challenges for practical deployment. Vision encoders are frequently deployed as standalone models on edge devices, where storage limits and inference overhead become major bottlenecks \cite{faghri2025mobileclip}.\footnote{To this end, several vendors provide deployment-ready lightweight vision encoders and toolchains for edge processors; see, e.g., \href{https://build.nvidia.com/nvidia/nvclip}{NVIDIA (NV-CLIP)}, \href{https://aihub.qualcomm.com/models/openai_clip}{Qualcomm AI Hub (OpenAI CLIP)}, and \href{https://github.com/hailo-ai/hailo-CLIP}{Hailo (Hailo-CLIP)}.} Vision encoders also serve as the visual backbones of many vision-language models (VLMs) \cite{liu2023visual,beyer2024paligemma}, where their computational cost remains substantial, especially for high-resolution images or video \cite{li2024llama,vasu2025fastvlm}. 
In video pipelines, the vision encoder can account for roughly 45\% of overall latency, and may even exceed the LLM prefilling cost at high resolutions \cite{li2024llama,vasu2025fastvlm}.

\begin{figure}[t]
    \vspace{+2.5em}
    \centering
    \includegraphics[width=0.98\columnwidth]{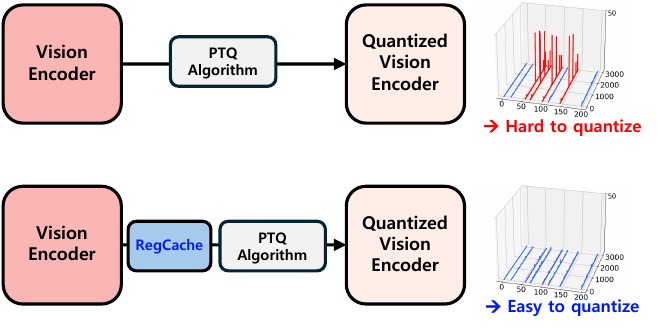}
    \caption{A diagram of the proposed RegCache framework, pre-processing the vision encoder before PTQ.}
    \label{fig:abstract_fig}
    \vspace{-1.5em}
\end{figure}

To address these challenges, post-training quantization (PTQ) offers a practical solution, reducing memory usage and computational cost without additional training \cite{choukroun2019low}. However, despite recent progress, PTQ methods for ViT-based vision encoders still suffer from significant performance degradation under low-bit quantization \cite{li2023repq,wu2025fima,zhong2025erq}. Moreover, unlike autoregressive large language models (LLMs), which are often memory-bound on GPU servers, vision encoders are typically non-autoregressive and thus more likely to operate in compute-bound regimes.\footnote{In contrast, the memory-bound nature of autoregressive LLM inference running on GPU servers renders weight-only quantization to be particularly effective.} Quantizing both activations and weights therefore allows high-precision matrix multiplications to be replaced with low-precision operations (e.g., INT8), hence reducing computational cost.

However, activation quantization of large pretrained transformers is challenging due to \textit{outlier} activations---i.e., a small number of extremely large activations that often arise in a few channels of later layers \cite{sun2024massive}. These outliers significantly expand the activation quantization range, leading to large quantization errors. While outlier-robust quantization has been widely studied for LLMs \cite{dettmers2022gpt3,xiao2023smoothquant,lin2024awq}, existing methods typically assign different precision levels or quantization ranges to individual tokens or channels. This introduces substantial overhead, making them difficult to apply in static activation quantization settings \cite{son2024prefixing,chen2024prefixquant}.

An emerging alternative is to directly mitigate outliers by prefixing \textit{attention sink} tokens---i.e., semantically meaningless tokens such as $\langle\mathtt{BOS}\rangle$ or $\langle\mathtt{SEP}\rangle$ that absorb large attention from other tokens \cite{xiao2024efficient,sun2024massive}. Recent studies on LLM quantization show that inserting the activations of these sink tokens as a prefix in each attention layer can dramatically reduce the activation magnitudes of other tokens, thereby improving PTQ performance \cite{yang2024mitigating,son2024prefixing,chen2024prefixquant}.

\begin{figure*}[t!]
    \centering
    \includegraphics[width=0.90\textwidth]{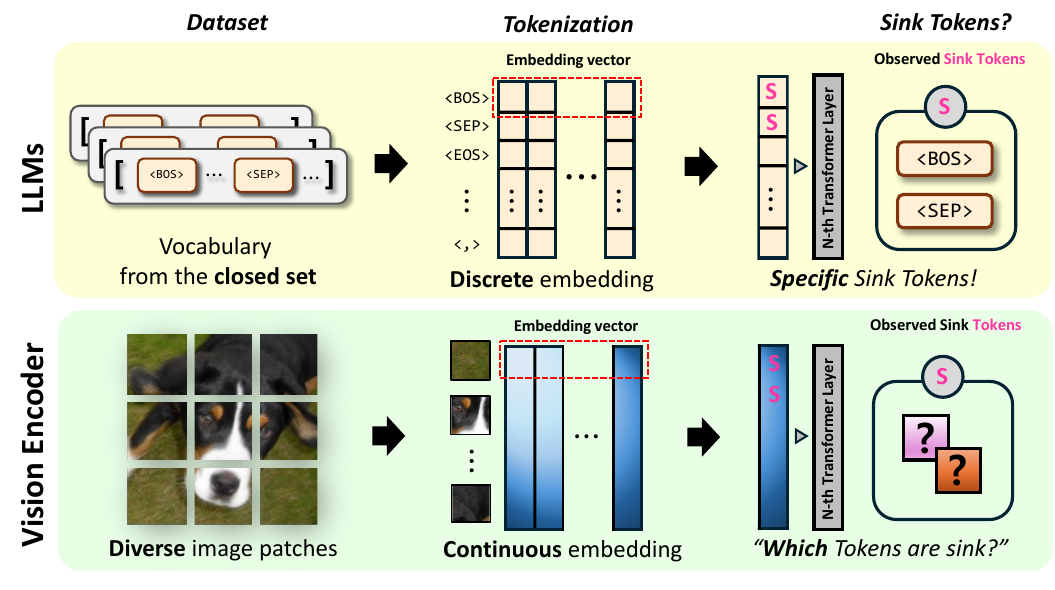}
    \caption{\textbf{Sink tokens in LLMs vs. vision encoders.} Unlike LLMs, where sink tokens exist in a closed-set vocabulary, vision encoders process diverse image patches continuously mapped into an embedding space, making sink token discovery more challenging.}
    \label{fig:placeholder}
\end{figure*}

Naturally, one may ask: \textit{Can we mitigate outliers in vision encoders by prefixing attention sinks}? Unfortunately, it remains unclear which token in a vision encoder (i.e., a patch embedding) could play a role analogous to attention sinks in language models. Unlike LLMs, typical vision encoders are not pretrained with tokens that are explicitly designed to be semantically meaningless (\cref{fig:placeholder}). Recent work suggests that introducing such meaningless tokens---termed \textit{registers}---during training can improve the interpretability of ViT-based models \cite{darcet2024vision}. However, incorporating registers remains uncommon in vision encoders.\footnote{In this regard, DINOv3 is a pleasant exception \cite{simeoni2025dinov3}.}

\paragraph{\textbf{Contribution.}} 
In this work, we introduce \textbf{\textit{RegCache}} (Register Caching), a novel prefix-based outlier mitigation algorithm for quantizing pretrained vision encoders. RegCache is inspired by the following empirical observation:
\begin{center}
\begin{tcolorbox}
   Sink tokens emerge gradually from the \textbf{middle} layers of vision encoders, giving rise to outliers. These tokens are \textbf{highly similar across images}, enabling their use as \textbf{reusable registers} for other inputs at test time.
\end{tcolorbox}
\end{center}
Based on these observations, which distinguish large pretrained vision encoders from relatively small ViTs, RegCache mitigates outliers by discovering and prefixing middle-layer registers in the form of a precomputed key-value (KV) cache. Notably, unlike in LLM prefixing \cite{son2024prefixing}, these tokens are inserted only in the middle-to-final layers and do not affect the early layers. RegCache also removes the remaining sink tokens that exhibit unusually large outliers. Through this process of adding and deleting tokens, RegCache replaces internally emerging sink tokens with external precomputed caches, preventing them from inflating the activation quantization range. Importantly, this procedure requires no additional training, rendering RegCache a versatile go-to method. In practice, RegCache can be integrated as a lightweight on-top module into existing PTQ pipelines.

Throughout experiments, we apply RegCache to diverse text- and self-supervised vision encoders and combine it with recent PTQ methods for ViTs, which can be further improved by mitigating large activation outliers. We observe that such combinations consistently improve the prediction accuracy of quantized vision encoders across all considered setups. Notably, these gains are most pronounced under extremely low-bit quantization (e.g., 4-bit).

\section{Related work}\label{sec:related_work}
\paragraph{\textbf{Outliers in large-scale transformers.}}
In large-scale transformers, some activations in certain layers can attain magnitudes significantly larger than others---a phenomenon known as the \textit{emergence of outliers} \cite{kovaleva2021bert,timkey2021all,bondarenko2021understanding,dettmers2022gpt3}. 
Sun et al. \cite{sun2024massive} provide a systematic analysis and show that these outliers arise from the softmax operation in self-attention in both LLMs and ViTs.
In LLMs, extreme activations are often concentrated on special tokens such as $\langle\mathtt{BOS}\rangle$ or $\langle\mathtt{SEP}\rangle$. In ViTs, several studies report that outlier tokens typically correspond to uninformative background patches, and removing them can improve internal representations \cite{darcet2024vision, jiang2025vision, lu2025artifacts}. However, it remains unclear which specific visual tokens consistently produce outliers in ViT-based vision encoders, as the corresponding patches vary across images. To address this gap, we observe that outlier tokens typically emerge in intermediate blocks and exhibit similar features across images and across a wide range of vision encoders, enabling them to be precomputed for downstream use cases such as PTQ.

\paragraph{\textbf{Improving vision transformers via controlling attention sink tokens.}}
Attention sinks, first highlighted by Xiao et al. \cite{xiao2024efficient}, are tokens with little or no semantic content that nonetheless attract excessive attention in both LLMs \cite{guo2024attention} and ViTs \cite{darcet2024vision}. In ViTs, these sink tokens act as noise in the attention map, hindering the model's ability to capture relations between patches and degrading downstream visual performance \cite{darcet2024vision, jiang2025vision, kang2025see, lu2025artifacts}.
To mitigate this issue, Darcet et al.~\cite{darcet2024vision} introduce an additional register token during training to absorb the attention sinks that emerge in image patches. More recently, Jiang et al.~\cite{jiang2025vision} propose a training-free \textit{test-time register} that relocates the maximum activations from channels that frequently exhibit outliers into an extra token during inference. RegCache, in contrast, is specifically tailored to the quantization scheme. We establish a formal correlation between quantization sensitivity and outlier emergence, and exploit this relationship to suppress the outliers that degrade PTQ performance. Furthermore, our method relies on pre-computed token-level operations that incur minimal overhead and require no per-image processing, which is efficient and well-aligned with the goal of quantization.

\paragraph{\textbf{Post-training quantization for vision transformers.}}
Prior work has explored reducing the inference cost of large-scale ViT-based models through PTQ. Early methods mitigate quantization errors by assigning dynamic bitwidths to self-attention-sensitive blocks \cite{liu2021post}. Subsequent studies attribute the low PTQ performance of ViTs to activation outliers arising from operations such as LayerNorm, softmax, and GELU. Motivated by this observation, RepQ-ViT \cite{li2023repq} and PTQ4ViT \cite{yuan2022ptq4vit} propose quantization schemes that isolate and minimize the impact of outliers. Other approaches address the heavy-tailed activation distribution; for example, NoisyQuant \cite{liu2023noisyquant} injects noise to reshape the activation distribution into a more quantization-friendly form. FIMA-Q \cite{wu2025fima} introduces round-function optimization for PTQ, while ERQ \cite{zhong2025erq} proposes a two-step procedure that sequentially quantizes activations and weights to reduce quantization error. Despite their methodological diversity, these approaches share a common premise: they are applied in the presence of outliers--whether through specialized quantizers for heavy-tailed activations \cite{yuan2022ptq4vit,zhong2025erq}, scale reparameterization \cite{li2023repq,zhong2025erq}, distribution reshaping \cite{liu2023noisyquant}, quantizer optimization \cite{wu2025fima}, or ridge regression \cite{zhong2025erq}. In contrast, RegCache directly suppresses outliers by leveraging a structural property of the vision encoder. As a result, RegCache can be easily integrated into existing PTQ pipelines to further reduce quantization error.

\paragraph{\textbf{Outlier mitigation by prefixing sink-like tokens.}}
Son et al.~\cite{son2024prefixing} mitigate outliers in LLMs by prefixing the sink-like tokens in the form of a KV cache. While RegCache inherits the high-level idea of prefixing register tokens, it introduces three technical innovations that account for the unique properties of ViTs. (1) RegCache constructs its register candidate set in a data-driven manner, whereas LLM methods simply use the dictionary, since visual tokens are continuous embeddings rather than entries of a fixed vocabulary. (2) While Son et al.~\cite{son2024prefixing} caches prefix tokens from the model input, where outliers arise in the early layers of LLMs, RegCache instead caches from the middle-layer, since ViT outliers emerge from the intermediate blocks (see~\ref{ssec:middle_layer_insperction}). (3) RegCache additionally removes tokens to eliminate leftover outliers, which is critical to performance (see Tab.~\ref{tab:ablation_study}).

\begin{figure*}[!t]
\begin{center}
\begin{minipage}{1.0\linewidth}
\centering
\includegraphics[width=1.0\linewidth]
{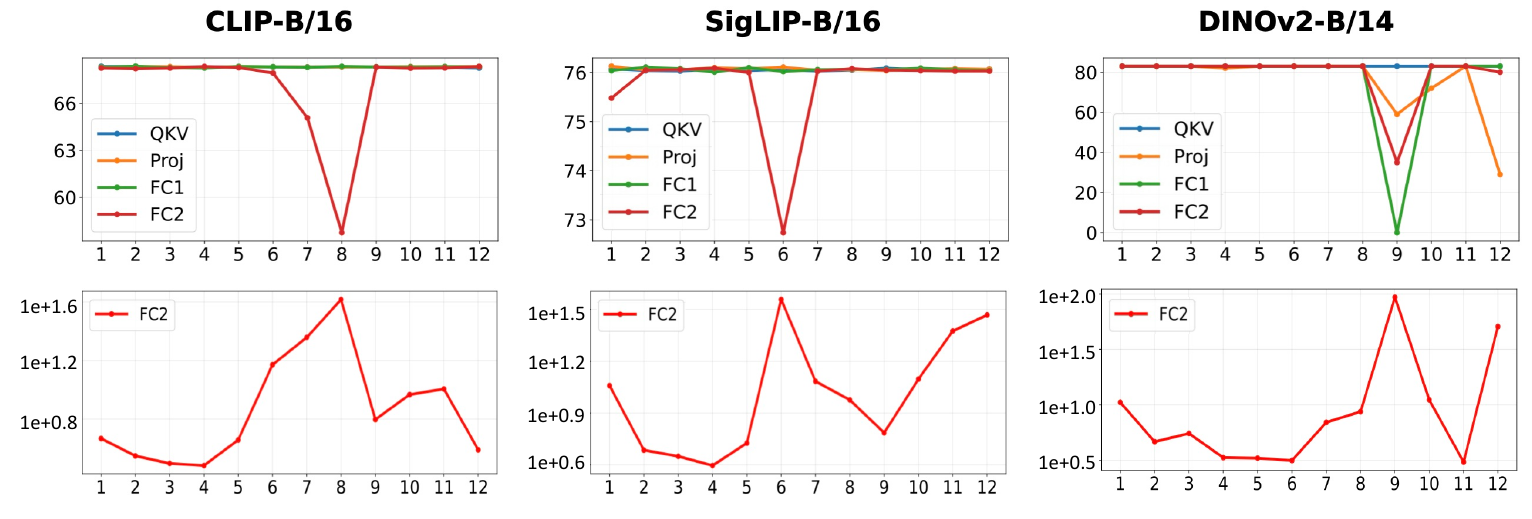}
\end{minipage}
\caption{\textbf{
(Top) Layerwise quantization sensitivity (\%).} We plot the zero-shot ImageNet-1k accuracy when quantizing each layer individually to W8A8. \textbf{(Bottom) Maximum norm of the FC2 layer input tokens for each layer.} We plot the largest $\ell_\infty$-norm across all tokens per image on a log scale, averaged over the ImageNet-1k validation set. For both plots, the x-axis denotes the transformer block index.}
\label{fig:quantization_sensitivity}
\end{center}
\end{figure*}
\section{A closer look at outliers in vision encoders}\label{sec:motivation}
Outliers in vision encoders tend to emerge at seemingly random background patch tokens, whereas in LLMs they typically occur at specific token positions or types \cite{sun2024massive,darcet2024vision}. This lack of consistency makes it difficult to mitigate outliers in vision encoders through prefixing \cite{son2024prefixing}, as such strategies require caching tokens that consistently induce outliers across arbitrary test-time inputs (i.e., registers).

In this section, we present three key findings that motivate an alternative strategy---\textit{identifying reusable sink tokens in the middle blocks of the model and prefixing them there}. These findings also shed light on why outliers in vision encoders predominantly emerge in the middle layers.

\begin{itemize}%[leftmargin=*,topsep=0pt,parsep=0pt]
\item \Cref{ssec:qsen}: Quantization-sensitive layers primarily occur in the middle of the network, where outliers emerge, while remaining low elsewhere; thus, prefixing is unnecessary in early layers.
\item \Cref{ssec:middle_layer_insperction}: We explain why outliers emerge later in the network: early layers first identify semantically meaningless tokens.
\item \Cref{ssec:cossim}: In the middle layers, outlier tokens become highly similar across images, suggesting that they act as reusable registers.
\end{itemize}

\subsection{Layerwise quantization sensitivity and outliers}
\label{ssec:qsen} 
We first analyze the quantization sensitivity of each layer in vision encoders and examine its connection to the emergence of activation outliers (i.e., FC2 inputs). 
In \cref{fig:quantization_sensitivity} (top), we report the layerwise quantization sensitivity, measured as the zero-shot ImageNet-1k accuracy when a given layer is quantized to W8A8. We observe that quantization-sensitive layers---i.e., layers that incur substantial accuracy drops when quantized---are localized to the MLP projection layers in one or two middle layers. In DINOv2, the performance degradation is particularly pronounced, and takes place in other layers as well. Moreover, as shown in \cref{fig:quantization_sensitivity} (bottom), these quantization-sensitive layers coincide with the blocks where activation outliers begin to emerge. Together, these suggest that outliers are a primary driver of performance degradation when quantizing vision encoders. In \cref{app:outlier_impact}, we further analyze the effect of outlier tokens under quantization, highlighting the importance of mitigating them to improve PTQ performance.

In particular, we establish a clear connection between quantized accuracy and the high-norm activation behavior. Our findings further suggest that monitoring FC2 activations or directly measuring quantization sensitivity can help identify the blocks where prefixing should be applied. Additional results for other vision encoders (OpenCLIP and SigLIP2) are provided in \cref{app:qsen}.

\subsection{Why the middle layers?}\label{ssec:middle_layer_insperction}
\begin{figure}[h]
    \centering
    \includegraphics[width=0.90\columnwidth]{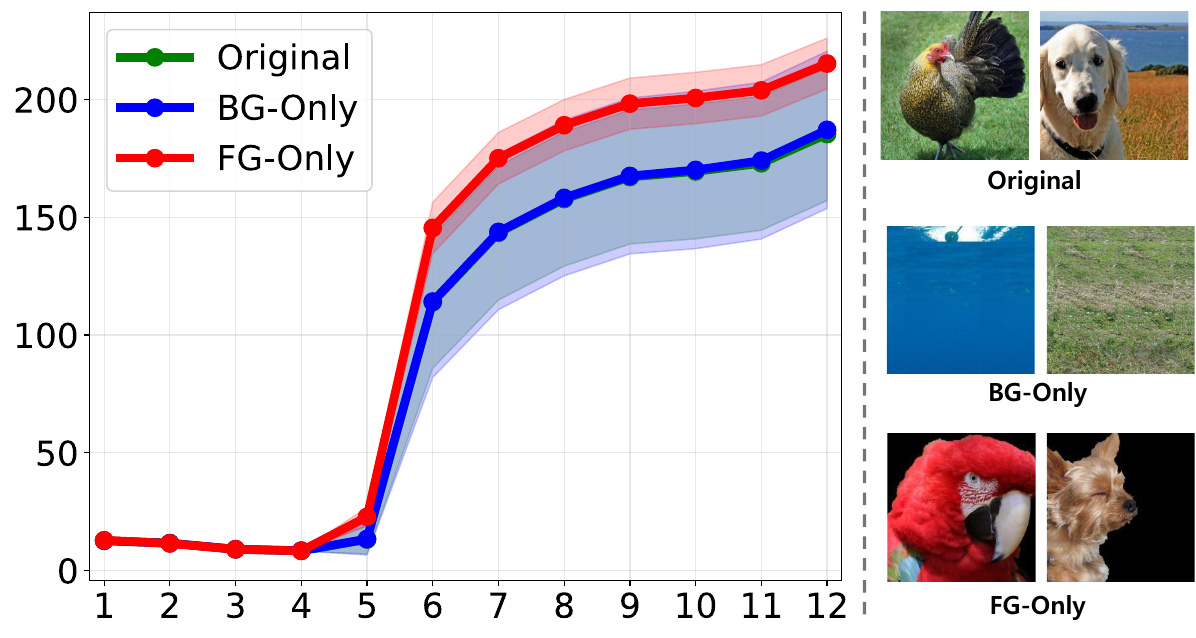}
    \caption{\textbf{Outliers in FG/BG-only images, for SigLIP-B/16 model.} Note that ``Original'' and ``BG-only'' nearly overlap.}
    \label{fig:bgfg}
    \vspace{-1.5em}
\end{figure}
A natural question arises: Why do outliers in vision encoders emerge in the middle layers, whereas in LLMs they appear in the early layers? We hypothesize that this difference stems from the difficulty of identifying \textit{semantically meaningless} tokens from raw image patches. Such tokens become distinguishable only after several processing blocks of the vision encoder. In contrast, in LLMs some tokens are clearly meaningless from the outset, such as $\langle\mathtt{BOS}\rangle,\:\langle\mathtt{SEP}\rangle$ (see \cref{fig:placeholder}, left).

To test this hypothesis, we design an experiment comparing outlier emergence in images where meaningless patches are easily identifiable against those where the distinction is less clear. Specifically, using the test set of ImageNet-9 \cite{xiao2021noise}, we compare foreground-only images---where the background pixels are zero-ed out---with the originals. In foreground-only images, semantically meaningless patches should be easier to identify, requiring fewer processing blocks.

As shown in \cref{fig:bgfg}, outliers in foreground-only images indeed emerge earlier and with larger magnitude than in the original images, supporting our hypothesis. In contrast, removing the foreground and retaining only the background leaves the outlier behavior largely unchanged (i.e., the curves nearly overlap). In \cref{app:freq}, we further analyze vision encoders trained with registers, where meaningless tokens are explicitly defined. These models exhibit behavior similar to LLMs, with outliers emerging from the early layers.

\subsection{Reusability of outlier tokens}\label{ssec:cossim}
\setlength{\columnsep}{5.5pt}
\begin{wraptable}{r}{0.5\columnwidth}
\vspace{-1.2em}
\centering
\caption{The average cosine similarity between two distinct groups of tokens in SigLIP-B/16.}
\small
\resizebox{0.95\linewidth}{!}{
\begin{tabular}{lc}
\toprule
\textbf{Token Type} & \textbf{Cosine sim.} \\
\midrule
Normal tokens & 0.26 {\scriptsize ($\pm$0.10)} \\
Outlier tokens & 0.89 {\scriptsize ($\pm$0.07)} \\
\bottomrule
\end{tabular}
\vspace{-2.0em}
}
\label{tab:cosine_similarity}
\end{wraptable}

Next, we examine the outlier tokens in the quantization-sensitive layer. Specifically, we measure the cosine similarity between middle-layer FC2 input outlier tokens (i.e., those with the largest $\ell_{\infty}$-norm) extracted from pairs of images. Using 64 randomly sampled images from the ImageNet-1k validation split, we compute the mean pairwise cosine similarity.

As shown in \cref{tab:cosine_similarity}, outlier tokens are highly similar across images, with a mean cosine similarity of $0.89$. In contrast, normal tokens are much less similar, with an average of only $0.26$. This suggests that outliers contain components largely independent of the input image and may represent shared features that persist across samples. In \cref{app:cossim}, we provide a theoretical explanation, showing that it arises from the large magnitude of the outliers and the alignment of their locations.

\begin{figure*}[!t]
    \centering
    \includegraphics[width=\textwidth]{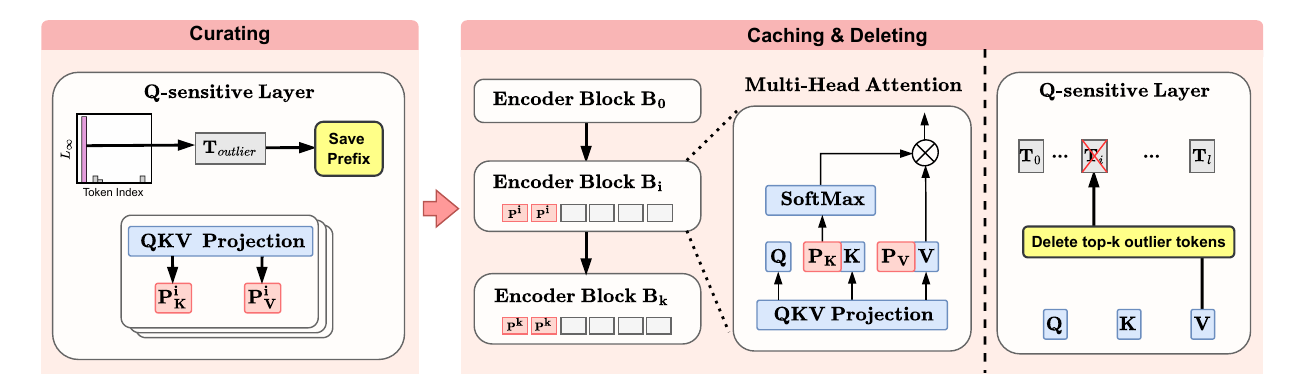}
\caption{\textbf{An overview of the proposed RegCache algorithm.} We identify a universal register by analyzing the inputs of quantization-sensitive layers across blocks. During inference, the register is inserted into each block, and outlier tokens are removed from the most quantization-sensitive layer.}
    \label{fig:method}
\end{figure*}

\section{Method}\label{sec:method}
Before introducing our method, we recall two observations from \cref{sec:motivation}:
\begin{itemize}%[leftmargin=*, itemsep=0pt, topsep=1pt]
\item In vision encoders, outliers tend to emerge in the middle layers (e.g., fully-connected layers in blocks 5--8), whereas in LLMs they begin to appear in the early layers.
\item Sink tokens (i.e., outlier-prone tokens) discovered in the middle layers are highly similar across images.
\end{itemize}
Combining these observations with prior findings in LLMs---prefixing additional sink tokens can mitigate outliers \cite{son2024prefixing}---we arrive at the following hypothesis:
\begin{center}
    \textit{``Middle-layer sink tokens from one image can act as registers and help mitigate outliers in vision encoders processing another image.''}
\end{center}

Based on this hypothesis, we propose RegCache (Register Caching), an outlier-mitigation algorithm that replaces internally emerging sink tokens by inserting register tokens discovered from reference images. In essence, RegCache operates in three steps (see \cref{fig:method}).

\newcommand{\inlineref}[1]{\textcolor{gray}{\footnotesize(\cref{#1})}}

\begin{enumerate}
\item \textbf{\textit{Curating}} a set of register candidate tokens---i.e., with large activation magnitudes---from a pool of reference images. \inlineref{ssec:curating}
\item \textbf{\textit{Caching}} the keys and values of selected register candidate tokens in the quantization-sensitive layers. \inlineref{ssec:caching}
\item \textbf{\textit{Deleting}} the sink tokens that have internally emerged---i.e., not inserted---to eliminate remaining outliers. \inlineref{ssec:deleting}
\end{enumerate}

RegCache requires no additional training and performs only a few rounds of validation on a reference task and dataset, as will be discussed below. As a result, the algorithm requires neither large amounts of data nor substantial computational resources. At inference time, RegCache temporarily adds and removes a few tokens, introducing only negligible overhead (see \cref{ssec:latency}).

\subsection{Curating}\label{ssec:curating}
Given a pretrained vision encoder, we first identify the quantization-sensitive layer of the model. We then construct a curated set of register candidate tokens by selecting the top-$k$ tokens with the largest activation at the input of this layer from a pool of reference images. 

\paragraph{\textbf{Identifying the quantization-sensitive layer.}} Following \cref{ssec:qsen}, we quantize each layer of the vision encoder independently, and select the one that yields the largest accuracy drop on a reference task as the quantization-sensitive layer. If a specific base quantization algorithm is intended for deployment, we use it during this analysis; otherwise, we use standard round-to-nearest quantization. As the reference task, we use ImageNet-1k classification (on the training split), which is widely regarded as representative of visual understanding. As a label-free alternative, we also provide a reconstruction-loss-based approach that leverages unlabeled data (see \cref{app:recon_loss}).

\paragraph{\textbf{Curating the set of register candidates.}} After identifying the quantization-sensitive layer, we construct a set of register candidate tokens likely to act as registers when inserted near this layer. Specifically, we run inference on a pool of reference images and select the tokens with the largest $\ell_{\infty}$ norm at the quantization-sensitive layer. Formally, let $l_{\mathrm{q}}$ denote the quantization-sensitive layer identified via the sensitivity analysis in \cref{ssec:qsen}, and let $\Phi_{l}(\mathbf{x})$ denote the set of tokens at the input of the $l$-th layer for an image $\mathbf{x}$. We construct the register candidate set as
\begin{align}
\mathcal{S} = \mathrm{argtop\textit{k}}\left\{\|\mathbf{z}\|_{\infty}~\big|~\mathbf{z}\in \Phi_{l_{q}}(\mathbf{x}), \quad \text{for some\:}\mathbf{x} \in \mathcal{I}_{\mathrm{ref}} \right\},
\end{align}
where $\mathcal{I}_{\mathrm{ref}}$ denotes the pool of reference images. In our experiments, we sample 50{,}000 images from the ImageNet-1k training split and set $k=100$. Since sink tokens often emerge several blocks before the quantization-sensitive layer, we perform the same search for up to three preceding blocks, constructing separate register candidate sets for each block.

\subsection{Caching}\label{ssec:caching}
Having the set of register candidates $\mathcal{S}$ constructed, we compose the register by simply averaging the KV caches of the register candidate tokens $\mathbf{z}^* \in \mathcal{S}$. We then search for the optimal $\tau^* \in \mathbb{N}$, the number of times the register is copied and inserted into the target vision encoder. Precisely, the search is done as follows.
\begin{itemize}
\item First, we compute KV caches for each register candidate token $\mathbf{z} \in \mathcal{S}$, for blocks starting from the few layers before the quantization-sensitive block, as well as all subsequent blocks using the \textit{unquantized} vision encoder.
\item Then, we insert the averaged KV cache to the \textit{quantized} vision encoder, with different numbers of copies. We vary $\tau$ within the range $\{1,2,\ldots,15\}$ and select $\tau^*$ as the one with the highest reference task accuracy.\footnote{The total search cost of our algorithm is about 1 hour when using a na\"ive RTN pseudo-quantization algorithm on an RTX 4090.} Here, as in \cref{ssec:curating}, we consider classification on the training split of ImageNet-1k dataset as our reference task. 
\end{itemize}

\subsection{Deleting}\label{ssec:deleting}
Finally, we apply a \textit{token deletion} process to the input of the
quantization-sensitive block (i.e., the block where $l_q$ is located)
in the vision encoder. At the inference phase, this process removes the sink tokens that emerge among the image patch tokens, thus removing any remaining outliers. Precisely, given some test image $\mathbf{x}_{\mathrm{test}}$, we select the tokens with the top-$\tilde{k}$ $\ell_{\infty}$ norm, \textit{i.e.},
\begin{align}
\mathcal{D} = \mathrm{argtop}\tilde{\mathrm{k}}\big\{\|\mathbf{z}\|_{\infty}~\big|~ \mathbf{z} \in \Phi_{l_q}(\mathbf{x}_{\mathrm{test}})\big\}
\end{align}
and remove these tokens. Here, similarly to the curating and caching steps, the number of tokens to be removed---\textit{i.e.}, $\tilde{k}$---is tuned using the reference task.
\begin{table*}[t]
\small
\centering
\caption{\textbf{Zero-shot classification accuracy on ImageNet-1k.} We have used various base quantization algorithms to quantize to 4/6/8 bits. The best results are marked in \textbf{bold}. Best/Average $\Delta$ denote the maximum/average accuracy gaps between each baseline and its RegCache counterpart, excluding the Na\"ive cases.}
\renewcommand{\arraystretch}{1.0}
\setlength{\tabcolsep}{5pt}
\resizebox{\linewidth}{!}{
\begin{tabular}{l ccc ccc ccc ccc ccc}
\toprule
& \multicolumn{3}{c}{\textbf{CLIP-B/16}} 
& \multicolumn{3}{c}{\textbf{OpenCLIP-B/16}} 
& \multicolumn{3}{c}{\textbf{SigLIP-B/16}} 
& \multicolumn{3}{c}{\textbf{SigLIP2-B/16}} 
& \multicolumn{3}{c}{\textbf{DINOv2-B/14}} \\
\cmidrule(lr){2-4} \cmidrule(lr){5-7} \cmidrule(lr){8-10} \cmidrule(lr){11-13} \cmidrule(lr){14-16}
\textbf{Method}
& \textbf{W4A4} & \textbf{W6A6} & \textbf{W8A8} 
& \textbf{W4A4} & \textbf{W6A6} & \textbf{W8A8} 
& \textbf{W4A4} & \textbf{W6A6} & \textbf{W8A8} 
& \textbf{W4A4} & \textbf{W6A6} & \textbf{W8A8} 
& \textbf{W4A4} & \textbf{W6A6} & \textbf{W8A8} \\
\midrule
FP32 
& \multicolumn{3}{c}{68.32}
& \multicolumn{3}{c}{70.22}
& \multicolumn{3}{c}{76.05}
& \multicolumn{3}{c}{78.47}
& \multicolumn{3}{c}{83.26} \\
\midrule
Na\"ive 
& 0.09 & 0.17 & 34.01
& 0.10 & 0.47 & 46.12
& 0.13 & 0.77 & 69.71
& 0.14 & 0.19 & 26.04
& 0.00 & 0.01 & 19.20 \\
\rowcolor{gray!10}
w/ RegCache 
& 0.07 & 0.44 & 59.71 
& 0.10 & 1.20 & 66.90 
& 0.10 & 24.41 & 74.38
& 0.11 & 0.26 & 72.35 
& 0.18 & 0.42 & 22.34 \\
\midrule
PTQ4ViT 
& 0.37 & 51.60 & 67.69
& 0.09 & 59.98 & 69.39
& 0.19 & 68.68 & 75.57 
& 0.28 & 41.54 & 76.92 
& 0.01 & 78.48 & \textbf{82.97} \\
\rowcolor{gray!10}
w/ RegCache 
& 1.78 & 55.49 & 67.86 
& 0.14 & 63.60 & 69.60 
& 0.59 & 72.38 & 75.75 
& 2.67 & 69.02 & 76.88 
& 2.38 & \textbf{78.55} & 82.92 \\
\midrule
RepQ-ViT 
& 1.83 & 53.25 & 67.39 
& 1.18 & 46.51 & 68.70 
& 21.36 & 73.32 & 75.23 
& 0.40 & 64.91 & 76.43 
& 4.52 & 19.61 & 82.27 \\
\rowcolor{gray!10}
w/ RegCache 
& 21.50 & 66.73 & \textbf{68.10} 
& 11.10 & 68.22 & 70.06 
& 33.76 & 74.52 & 75.94 
& 10.15 & 75.40 & 77.13 
& 4.97 & 22.38 & 81.55 \\
\midrule
NoisyQuant 
& 0.34 & 46.19 & 63.20 
& 2.50 & 59.05 & 67.08 
& 1.78 & 71.10 & 75.50 
& 0.46 & 44.50 & 70.83 
& 1.61 & 49.25 & 71.46 \\
\rowcolor{gray!10}
w/ RegCache 
& 1.42 & 57.41 & 65.82 
& 10.41 & 67.51 & 69.36 
& 9.19 & 72.28 & 75.64 
& 0.54 & 62.60 & 76.44 
& 2.04 & 48.65 & 70.38 \\
\midrule
FIMA-Q 
& 50.41 & 66.51 & 67.63 
& 60.09 & 67.24 & 68.53 
& 76.05 & 76.06 & 76.06 
& \textbf{78.48} & \textbf{78.47} & \textbf{78.48} 
& OOM & OOM & OOM \\
\rowcolor{gray!10}
w/ RegCache 
& \textbf{62.08} & 66.70 & 67.86 
& \textbf{65.11} & 68.62 & 69.13 
& \textbf{76.13} & \textbf{76.12} & \textbf{76.12} 
& 78.38 & 78.37 & 78.37
& OOM & OOM & OOM \\
\midrule
ERQ 
& 1.56 & 39.64 & 67.99 
& 43.15 & 66.70 & 69.25 
& 57.76 & 74.44 & 75.95 
& 10.56 & 74.75 & 78.05 
& 6.53 & 78.03 & 82.59 \\
\rowcolor{gray!10}
w/ RegCache 
& 46.07 & \textbf{66.79} & 68.09
& 54.76 & \textbf{69.09} & \textbf{70.12} 
& 60.99 & 75.25 & 75.98 
& 28.98 & 75.32 & 78.45 
& \textbf{9.15} & 77.88 & 82.54 \\
\midrule
Best $\Delta$
& \incr{+44.51} & \incr{+27.15} & \incr{+2.62}
& \incr{+11.61} & \incr{+21.84} & \incr{+2.28}
& \incr{+12.40} & \incr{+3.38} & \incr{+0.71}
& \incr{+18.42} & \incr{+27.48} & \incr{+5.61}
& \incr{+2.62} & \incr{+2.77} & \decr{-0.05} \\
Average $\Delta$
& \incr{+15.67} & \incr{+11.19} & \incr{+0.77}
& \incr{+6.90} & \incr{+7.53} & \incr{+1.06}
& \incr{+4.70} & \incr{+1.35} & \incr{+0.25}
& \incr{+6.11} & \incr{+11.31} & \incr{+1.36}
& \incr{1.11} & \incr{+0.52} & \decr{-0.47} \\
\bottomrule
\end{tabular}
}
\label{tab:naive_best_config_with_diff}
\end{table*} 
\begin{table*}[ht!]
\small
\centering
\caption{\textbf{Zero-shot image--text retrieval performance of CLIP and SigLIP on MS-COCO (R@1).} The results are reported in the same format as in \cref{tab:naive_best_config_with_diff}.}
\renewcommand{\arraystretch}{1.15}
\setlength{\tabcolsep}{4pt}
\resizebox{0.8\linewidth}{!}{
\begin{tabular}{lcccccccccccc}
\toprule
& \multicolumn{6}{c}{\textbf{CLIP-B/16}} & \multicolumn{6}{c}{\textbf{SigLIP-B/16}} \\
\cmidrule(lr){2-7}\cmidrule(lr){8-13}
& \multicolumn{3}{c}{\textbf{Image $\rightarrow$ Text}} & \multicolumn{3}{c}{\textbf{Text $\rightarrow$ Image}}
& \multicolumn{3}{c}{\textbf{Image $\rightarrow$ Text}} & \multicolumn{3}{c}{\textbf{Text $\rightarrow$ Image}} \\
\cmidrule(lr){2-4}\cmidrule(lr){5-7}\cmidrule(lr){8-10}\cmidrule(lr){11-13}
& W4A4 & W6A6 & W8A8
& W4A4 & W6A6 & W8A8
& W4A4 & W6A6 & W8A8
& W4A4 & W6A6 & W8A8 \\
\midrule

\multicolumn{1}{l}{FP32}
& 52.94 & 52.94 & 52.94
& 32.73 & 32.73 & 32.73
& 67.68 & 67.68 & 67.68
& 47.19 & 47.19 & 47.19 \\
\midrule

Na\"ive
& 0.00 & 0.00 & 22.76
& 0.01 & 0.06 & 14.08
& 0.00 & 0.34 & 60.04
& 0.04 & 0.86 & 41.80 \\
\rowcolor{gray!10}
w/ RegCache
& 0.02 & 0.22 & 46.10
& 0.04 & 0.44 & 28.02
& 0.04 & 12.06 & 65.76
& 0.04 & 12.70 & 46.30 \\
\midrule

PTQ4ViT
& 0.06 & 37.28 & 52.78
& 0.17 & 23.00 & 32.00
& 0.20 & 60.66 & 66.86
& 0.78 & 41.73 & 47.16 \\
\rowcolor{gray!10}
w/ RegCache
& 0.52 & 43.46 & 53.54
& 0.14 & 26.51 & 32.48
& 2.10 & 61.42 & 67.72
& 1.78 & 43.05 & 47.61 \\
\midrule

RepQ-ViT
& 0.32 & 29.06 & 44.52
& 0.47 & 15.90 & 23.01
& 5.14 & 37.50 & 65.90
& 6.21 & 26.41 & 46.33 \\
\rowcolor{gray!10}
w/ RegCache
& 4.32 & 38.68 & 45.58
& 3.37 & 19.70 & 23.68
& 14.92 & 61.06 & 66.12
& 9.37 & 43.63 & 46.42 \\
\midrule

NoisyQuant
& 0.12 & 25.26 & 48.94
& 0.19 & 18.02 & 31.07
& 0.29 & 52.52 & 67.10
& 1.18 & 33.36 & 46.76 \\
\rowcolor{gray!10}
w/ RegCache
& 0.37 & 33.86 & 49.10
& 0.71 & 22.28 & 30.40
& 2.53 & 63.28 & 67.24
& 3.81 & 43.25 & \textbf{47.64} \\
\midrule

FIMA-Q
& 40.92 & 52.90 & 53.58
& 27.71 & 32.33 & 32.76
& 67.66 & 67.70 & 67.62
& 47.23 & 47.22 & 47.21 \\
\rowcolor{gray!10}
w/ RegCache
& \textbf{51.82} & \textbf{53.22} & \textbf{53.66}
& \textbf{33.43} & \textbf{32.88} & \textbf{33.01}
& \textbf{68.14} & \textbf{68.14} & \textbf{68.18}
& \textbf{47.64} & \textbf{47.62} & 47.62 \\
\midrule

ERQ
& 0.32 & 23.62 & 44.78
& 3.71 & 15.58 & 23.31
& 28.96 & 61.94 & 65.80
& 20.70 & 43.48 & 46.46 \\
\rowcolor{gray!10}
w/ RegCache
& 13.78 & 40.88 & 44.86
& 6.52 & 19.82 & 23.52
& 32.66 & 62.30 & 66.16
& 23.21 & 44.61 & 46.63 \\
\midrule

Best $\Delta$
& \incr{+13.46} & \incr{+17.26} & \incr{+1.06}
& \incr{+5.72} & \incr{+4.26} & \incr{+0.67}
& \incr{+9.78} & \incr{+23.56} & \incr{+0.86}
& \incr{+5.41} & \incr{+17.22} & \incr{+0.88} \\
Average $\Delta$
& \incr{+5.81} & \incr{+8.40} & \incr{+0.43}
& \incr{+2.19} & \incr{+3.27} & \incr{+0.19}
& \incr{+3.55} & \incr{+7.18} & \incr{+0.43}
& \incr{+3.10} & \incr{+5.99} & \incr{+0.40} \\
\bottomrule
\end{tabular}
}
\label{tab:retrieval_main}
\end{table*}

\section{Experiments}\label{sec:result}
In this section, we first describe our experimental setup in \cref{ssec:setup}. Next, we report results for both standalone vision encoders and VLMs across various benchmarks in \cref{ssec:main_results}, demonstrating consistent gains from RegCache over baselines including low-bit regimes. We also assess the generalizability of the precomputed prefix tokens and further validate our design through analyses of prefix-token behavior, ablation studies, and latency measurements.

\subsection{Setup}\label{ssec:setup}
\paragraph{\textbf{Standalone vision encoders.}} We evaluate the proposed method on five widely used vision encoders: (1) CLIP \cite{radford2021learning}, (2) OpenCLIP \cite{cherti2023openclip}, (3) SigLIP \cite{zhai2023siglip}, (4) SigLIP2 \cite{tschannen2025siglip2}, and (5) DINOv2 \cite{oquab2024dinov}, which cover diverse training configurations. For evaluation, we measure the quality of the quantized vision encoders using zero-shot accuracy on two downstream tasks: (1) Image classification on ImageNet-1k \cite{deng2009imagenet} and (2) text-image retrieval on MS-COCO \cite{lin2014mscoco}. 
To further validate the generalizability of the prefix searched from ImageNet-1k, we evaluate on diverse classification benchmarks, including Stanford Cars \cite{stanfordcar}, Flowers-102 \cite{flowers}, Food-101 \cite{food}, CIFAR-100, Caltech101 \cite{fei2004caltech101}, Oxford-IIIT Pet \cite{parkhi12oxfordpet}, and DTD \cite{cimpoi14dtd}.

\begin{table*}[t]
\small
\centering
\renewcommand{\arraystretch}{1.2}
\caption{Zero-shot classification accuracy (\%) on various datasets.}
\label{tab:generalizability}
\setlength{\tabcolsep}{4pt}
\resizebox{\textwidth}{!}{
\begin{tabular}{llccccccc}
\toprule
\textbf{Model} & \textbf{Method} 
& \textbf{StanfordCars} & \textbf{Flowers-102} & \textbf{Food-101} & \textbf{CIFAR-100}
& \textbf{Caltech101} & \textbf{OxfordIIIPet} & \textbf{DTD} \\
\midrule
\multirow{3}{*}{CLIP-B/16} 
& FP32       & 64.41 & 65.88 & 85.22 & 68.44 & 85.82 & 88.03 & 44.31 \\
& Na\"ive     & 29.76 & 26.20 & 33.30 & 35.96 & 70.59 & 74.19 & 30.05 \\
& \cellcolor{gray!10} w/ RegCache 
& \cellcolor{gray!10} 49.96 \inc{+20.20} 
& \cellcolor{gray!10} 55.39 \inc{+29.19} 
& \cellcolor{gray!10} 74.68 \inc{+41.38} 
& \cellcolor{gray!10} 51.87 \inc{+15.91}
& \cellcolor{gray!10} 83.22 \inc{+12.63}
& \cellcolor{gray!10} 86.21 \inc{+12.02}
& \cellcolor{gray!10} 42.82 \inc{+12.77} \\
\midrule
\multirow{3}{*}{OpenCLIP-B/16} 
& FP32       & 88.07 & 69.88 & 83.77 & 76.82 & 88.37 & 88.74 & 54.89 \\
& Na\"ive     & 74.85 & 42.97 & 36.44 & 40.61 & 75.44 & 78.09 & 36.28 \\
& \cellcolor{gray!10} w/ RegCache 
& \cellcolor{gray!10} 85.85 \inc{+11.00} 
& \cellcolor{gray!10} 68.06 \inc{+25.09} 
& \cellcolor{gray!10} 80.80 \inc{+44.36} 
& \cellcolor{gray!10} 71.73 \inc{+31.12}
& \cellcolor{gray!10} 86.88 \inc{+11.44}
& \cellcolor{gray!10} 87.00 \inc{+8.91}
& \cellcolor{gray!10} 51.70 \inc{+15.42} \\
\midrule
\multirow{3}{*}{SigLIP-B/16} 
& FP32       & 90.81 & 82.63 & 89.34 & 72.33 & 93.10 & 92.15 & 63.51 \\
& Na\"ive     & 87.97 & 75.26 & 78.31 & 54.79 & 92.65 & 90.24 & 62.82 \\
& \cellcolor{gray!10} w/ RegCache 
& \cellcolor{gray!10} 89.73 \inc{+1.76} 
& \cellcolor{gray!10} 80.32 \inc{+5.06} 
& \cellcolor{gray!10} 88.17 \inc{+9.86} 
& \cellcolor{gray!10} 66.87 \inc{+12.08}
& \cellcolor{gray!10} 92.65 \inc{+0.00}
& \cellcolor{gray!10} 91.33 \inc{+1.09}
& \cellcolor{gray!10} 63.62 \inc{+0.80} \\
\midrule
\multirow{3}{*}{SigLIP2-B/16} 
& FP32       & 92.74 & 83.38 & 90.65 & 77.10 & 93.19 & 92.31 & 61.33 \\
& Na\"ive     & 35.12 & 26.38 & 30.55 & 20.92 & 67.48 & 58.22 & 37.23 \\
& \cellcolor{gray!10} w/ RegCache 
& \cellcolor{gray!10} 88.20 \inc{+53.08} 
& \cellcolor{gray!10} 76.50 \inc{+50.12} 
& \cellcolor{gray!10} 86.47 \inc{+55.92} 
& \cellcolor{gray!10} 59.78 \inc{+38.86}
& \cellcolor{gray!10} 92.26 \inc{+24.78}
& \cellcolor{gray!10} 89.70 \inc{+31.48}
& \cellcolor{gray!10} 57.77 \inc{+20.54} \\
\bottomrule
\end{tabular}
}
\end{table*} 
\begin{table*}[t!]
    \centering
    \setlength{\tabcolsep}{6pt}
    \renewcommand{\arraystretch}{1.15}
    \caption{\textbf{VLM performance.}
    We evaluate our method on image and video benchmarks using Qwen3-VL-\{2B,8B\} models under 4-bit quantization.}
    \footnotetext[1]{We follow the setup in \href{https://docs.nvidia.com/deeplearning/triton-inference-server/user-guide/docs/tutorials/Popular_Models_Guide/Llava1.5/llava_trtllm_guide.html}{NVIDIA Triton / LLaVA-TRTLLM guide}.}
    \label{tab:vlm_performance}

    \resizebox{\linewidth}{!}{%
    \begin{tabular}{lcccccccc}
    \toprule
    \multirow{2}{*}{\textbf{Method}} &
    \multicolumn{4}{c}{\textbf{Qwen3-VL-2B}} &
    \multicolumn{4}{c}{\textbf{Qwen3-VL-8B}} \\
    \cmidrule(lr){2-5}\cmidrule(lr){6-9}
    &
    \multicolumn{2}{c}{\textbf{Image}} &
    \multicolumn{2}{c}{\textbf{Video}} &
    \multicolumn{2}{c}{\textbf{Image}} &
    \multicolumn{2}{c}{\textbf{Video}} \\
    \cmidrule(lr){2-3}\cmidrule(lr){4-5}\cmidrule(lr){6-7}\cmidrule(lr){8-9}
    &
    \textbf{GQA} & \textbf{VQAv2} & \textbf{Video-MME} & \textbf{MLVU} &
    \textbf{GQA} & \textbf{VQAv2} & \textbf{Video-MME} & \textbf{MLVU} \\
    \midrule

    Full-precision
        & 58.59 & 77.05 & 73.44 & 68.34
        & 59.04 & 77.86 & 78.78 & 78.19 \\
    Na\"ive
        & 27.12 & 10.86 & 32.11 & 37.40
        & 22.32 & 26.14 & 35.33 & 39.33 \\
    RepQ-ViT
        & 38.77 & 32.03 & 44.56 & 41.17
        & 37.84 & 46.29 & 50.44 & 44.89 \\
    w/ RegCache
        & 42.11 \incr{(+3.34)} & 42.72 \incr{(+10.69)} & 49.22 \incr{(+4.66)} & 43.47 \incr{(+2.30)}
        & 44.00 \incr{(+6.16)} & 50.44 \incr{(+4.15)} & 52.89 \incr{(+2.45)} & 45.40 \incr{(+0.51)} \\
    \bottomrule
    \end{tabular}%
    }
\end{table*}

\paragraph{\textbf{Vision-language models.}} We primarily use Qwen3-VL \cite{bai2025qwen3} (2B and 8B) as a strong, widely adopted open-source VLM that can handle both image and video. Furthermore, to demonstrate the broad applicability of our method, we also evaluate it on other VLMs, such as LLaVA \cite{liu2023visual} (see \cref{app:app_vlm}).

We evaluate image understanding on GQA \cite{hudson2018gqa} and VQAv2 \cite{goyal2017vqav2}, as visual question-answering benchmarks. For video understanding, we use Video-MME \cite{fu2025videomme} to assess multi-domain reasoning and MLVU \cite{zhou2025mlvu}, which includes long-video evaluation where vision-encoder overhead can become a critical bottleneck.

\paragraph{\textbf{Base quantization algorithms and details.}}
To assess the broad applicability of RegCache as an effective on-top method, we evaluate it on four distinct categories of ViT PTQ baselines: (1) scaling factor optimization (PTQ4ViT \cite{yuan2022ptq4vit}, RepQ-ViT \cite{li2023repq}, and FIMA-Q \cite{wu2025fima}), (2) rounding function optimization (ERQ \cite{zhong2025erq} and FIMA-Q \cite{wu2025fima}), (3) weight correction (ERQ \cite{zhong2025erq}), and (4) activation distribution shaping (NoisyQuant \cite{liu2023noisyquant}) to reduce quantization error.

Additionally, for CLIP and SigLIP models, prefixes are inserted from the searched layer to the final layer. DINOv2, trained in a self-supervised manner, exhibits different behavior compared to CLIP and SigLIP models; consequently, we find that inserting the prefix only at the searched layer yields better results.

Further details on the experimental setup can be found in \cref{app:app_baselines}.

\subsection{Main results}\label{ssec:main_results}
\paragraph{\textbf{Standalone vision encoders.}} In \cref{tab:naive_best_config_with_diff}, we report the zero-shot image classification accuracy on ImageNet-1k dataset. We observe that the baselines combined with RegCache consistently achieve better accuracy in most settings. Specifically, baselines with RegCache outperform the base quantization methods in terms of both best accuracy gap (Best~$\Delta$) and average accuracy gap (Average~$\Delta$). Only one setup---DINOv2---exhibits a negligible accuracy drop.

For zero-shot image–text retrieval (\cref{tab:retrieval_main}), we observe that RegCache consistently improves performance across all setups. These results indicate RegCache can integrate well with other quantization methods across diverse tasks.

\begin{table*}[h!]
    \centering
    \caption{\textbf{Latency results.} We measure the latency (ms) of each model on a single NVIDIA A6000 GPU using TensorRT \cite{trt, trtll}. (a) Standalone vision encoder latency for CLIP-B/16 and SigLIP-B/16. (b) Vision encoder (VE) and language model (LM) prefill latency on Qwen3-VL-2B. We also report the end-to-end speedup (Accel.) achieved by quantizing only the vision encoder, while the LM is quantized to INT8.}
    \label{tab:latency_results}
    \setlength{\tabcolsep}{6pt}
    \renewcommand{\arraystretch}{1.1}
    % ===================== (a) Standalone vision encoder =====================
    \begin{subtable}[t]{0.44\linewidth}
        \centering
        \caption{\textbf{Standalone VE.} Batch size: 64.}
        \label{tab:latency_st_ve}
        \resizebox{\linewidth}{!}{%
        \begin{tabular}{@{}llcc@{}}
        \toprule
        \textbf{Model} & \textbf{Method} & \textbf{Latency} (ms) & \textbf{Accel.} \\
        \midrule

        \multirow{3}{*}{\textbf{CLIP-B/16}}
          & Full-precision    & 132.13 & -- \\
          & INT8              & 60.64  & 2.18$\times$ \\
          & INT8 + RegCache   & 61.27 \textcolor{gray}{\scriptsize ({+1.04\%})} & 2.16$\times$ \\
        \midrule

        \multirow{3}{*}{\textbf{SigLIP-B/16}}
          & Full-precision    & 128.49 & -- \\
          & INT8              & 62.29  & 2.06$\times$ \\
          & INT8 + RegCache   & 63.25 \textcolor{gray}{\scriptsize ({+1.54\%})} & 2.03$\times$ \\

        \bottomrule
        \end{tabular}%
        }
    \end{subtable}
    \hfill
    % ===================== (b) Acceleration in VLM =====================
    \begin{subtable}[t]{0.524\linewidth}
        \centering
        \caption{\textbf{Qwen3-VL-2B}}
        \label{tab:latency_vlm}
        \resizebox{\linewidth}{!}{%
        \begin{tabular}{@{}l l cc c@{}}
        \toprule
        \textbf{Data} & \textbf{Method} & \textbf{VE} (ms) & \textbf{LM Prefill} (ms) & \textbf{Accel.} \\
        \midrule
        \multirow{3}{*}{\textbf{Image}}
          & Full-precision    & 30.21 & 51.76 & -- \\
          & INT8              & 8.33 & 11.63 & 2.10$\times$ \\
          & INT8 + RegCache   & 8.40 \textcolor{gray}{\scriptsize ({+0.80\%})} & 11.81 & 2.08$\times$ \\
        \midrule
        \multirow{3}{*}{\textbf{Video}}
          & Full-precision    & 79.18 & 124.00 & -- \\
          & INT8              & 10.43 & 27.91 & 2.79$\times$ \\
          & INT8 + RegCache   & 10.44 \textcolor{gray}{\scriptsize ({+0.01\%})} & 27.98 & 2.79$\times$ \\
        \bottomrule
        \end{tabular}%
        }
    \end{subtable}
\end{table*}

In particular, RegCache benefits low-precision quantization (i.e., 4-bit and 6-bit) most, where activation outliers affect quantization more severely. Conversely, the gain shrinks when headroom is limited, e.g., at high bitwidths with advanced base PTQ methods that already compensate for quantization errors. The gain is also larger on softmax-based models (e.g., CLIP) than on SigLIP, consistent with the common understanding that softmax is a key contributor to outliers.

\paragraph{\textbf{Vision-language models.}}
We further evaluate our quantized vision encoders within VLMs on image and video benchmarks (\cref{tab:vlm_performance}). All results are reported in the 4-bit setting on top of the PTQ baseline (RepQ-ViT), reflecting a highly constrained deployment setting. As shown in \cref{tab:vlm_performance}, RegCache consistently improves accuracy in VLMs, demonstrating robust gains across benchmarks.

\paragraph{\textbf{Generalizability of prefixes.}}
Since the prefix search procedure in RegCache involves validation on the training split of the ImageNet-1k dataset, we additionally assess whether the learned prefixes generalize to other datasets, as the register token might have overfit to ImageNet-1k. In this spirit, we perform zero-shot classification on other datasets, with the results reported in \cref{tab:generalizability}. 
These results indicate that the prefix learned on ImageNet-1k remains effective on other datasets, suggesting that it acts as a transferable register token.

\subsection{Analyses}
\paragraph{\textbf{Outlier reduction.}}
\cref{tab:outlier_control} illustrates the change in the maximum token norm of the quantization-sensitive layer input when RegCache is applied. As expected, the maximum token norm decreases: This reduction effectively narrows the dynamic range of quantization, thereby improving quantization performance. 

\begin{table}[t]
\centering
\setlength{\tabcolsep}{20pt}
\caption{\textbf{Reduction in maximum token norm within the input of quantization-sensitive layers in W8A8.}
We report the mean across 500 image samples.}\label{tab:outlier_control}
\resizebox{0.7\columnwidth}{!}{
\begin{tabular}{@{}lcc@{}}
\toprule
\multicolumn{1}{c}{\multirow{2}{*}{\textbf{Model}}} &
\multicolumn{2}{c}{\textbf{Max token}} \\
\cmidrule(l{6pt}r{6pt}){2-3}
\multicolumn{1}{c}{} &
\textbf{Vanilla} & \textbf{w/ RegCache} \\
\midrule
CLIP     & 41.38 & 11.45 \\
OpenCLIP & 92.78 & 9.64 \\
SigLIP   & 35.82 & 3.64 \\
SigLIP2  & 148.20 & 15.16 \\
\bottomrule
\end{tabular}}
\end{table}

\begin{table}[t]
\centering
\caption{\textbf{Ablation studies.} We compare the contribution of each component of RegCache on SigLIP-B/16 and SigLIP2-B/16. Caching and deleting play a complementary role, enabling a wide coverage over a wide range of vision encoders.}
\vspace{0.5em}
\label{tab:ablation_study}
\setlength{\tabcolsep}{8pt}
\resizebox{1\columnwidth}{!}{
\begin{tabular}{lcc}
\toprule
\textbf{Method} & \textbf{SigLIP-B/16} & \textbf{SigLIP2-B/16} \\
\midrule
Baseline                         & 69.71 & 26.04 \\
Prefix Caching                   & 74.37 & 23.82 \\
Token Deleting                   & 42.41 & 69.06 \\
\midrule
Prefix Caching + Token Deleting  & \textbf{74.38} & \textbf{72.35} \\
\bottomrule
\end{tabular}
}
\end{table}

\paragraph{\textbf{Ablation.}}
To support our design choices, we ablate two components of RegCache: (1) prefix caching and (2) token deleting. In \cref{tab:ablation_study}, we present the ablation study results for SigLIP and SigLIP2.
The results show that the two stages, prefix caching and token deleting, act synergistically, yielding the best performance when both components are used together for both vision encoders. Surprisingly, when only one of the two steps is applied, we obtain even worse results than in the na\"ive case, further supporting the validity of our design choice. 

\subsection{Latency}\label{ssec:latency}
We measure latency for both standalone vision encoders (\cref{tab:latency_st_ve}) and VLMs (\cref{tab:latency_vlm}), and find that RegCache adds only marginal overhead (up to 1.54\%). For VLMs, we profile time-to-first-token (TTFT): quantizing the vision encoder yields over a 2$\times$ TTFT reduction with the LLM quantized to INT8, indicating that the vision encoder becomes the primary bottleneck once the LLM is accelerated. We further observe that this acceleration is even more pronounced for video inputs, where processing multiple frames incurs substantial compute.

\subsection{Other experiments}
Beyond the main experiments, we conduct additional experiments as follows presented in the supplementary materials: 
\begin{itemize}%[leftmargin=*,topsep=0pt,parsep=0pt,itemsep=0pt]
    \item Retrieval results with other vision encoders \hfill {\color{gray}$\triangleright$ \cref{app:app_rtrv}}
    \item Visualization of searched register tokens \hfill {\color{gray}$\triangleright$ \cref{app:app_vizreg}}
    \item Token deletion effect analysis with segmentation task
    \hspace*{\fill}{\color{gray}$\triangleright$ \cref{app:segmentation}}
    \item Hyperparameter-sensitivity analysis
    \hspace*{\fill}{\color{gray}$\triangleright$ \cref{app:hp_analysis}}
    \item RegCache using reconstruction loss \hfill
    {\color{gray}$\triangleright$ \cref{app:recon_loss}}
    \item Using FP16 instead of token deleting \hfill
    {\color{gray}$\triangleright$ \cref{app:fp16token}}
    \item Combining with weight-only quantization \hfill {\color{gray}$\triangleright$ \cref{app:app_woq}}
    \item Combining with Hadamard rotation algorithms \hfill {\color{gray}$\triangleright$ \cref{app:hadamard}}
    \item Combining with LLM outlier-mitigation methods \hfill {\color{gray}$\triangleright$ \cref{app:SmoothQuant}}
\end{itemize}

\section{Conclusion}\label{sec:conclusion}
We introduce a training-free outlier mitigation algorithm, \textit{RegCache}. RegCache can serve as an on-top method combined with existing PTQ algorithms for large-scale transformer-based vision encoders. Through extensive experiments, we demonstrate that RegCache consistently improves quantization performance across various tasks, indicating that it is indeed synergistic with other quantization methods. Our analyses reveal that RegCache suppresses activation outliers in quantization-sensitive layers, thereby narrowing the dynamic range of the input and improving quantization performance. Furthermore, we take a step toward identifying register tokens that are optimal for the quantization of vision encoders---a task that is inherently more elusive than for language models.

\paragraph{\textbf{Limitations.}} A major limitation of our method is the need to tune several additional hyperparameters, such as the maximum number of tokens to delete and the number of prefix tokens.
Another limitation is that the number of prefix and deleted tokens must be selected heuristically for each vision encoder and base quantization algorithm.

\paragraph{\textbf{Discussions and future directions.}} Our work covers a variety of vision encoders, including those trained on multimodal data (e.g., CLIP) and those trained on vision-only data (e.g., DINOv2). In our experiments, we observe that quantization-related measures (e.g., quantization sensitivity) behave somewhat differently across these cases, warranting further study. Another research direction arises from the differences between LLMs and ViT-based vision encoders: their outlier behavior differs significantly (see \cref{app:freq} for an extended discussion). Understanding this phenomenon would benefit a wide range of domains, including quantization and representation learning. 

\section*{Acknowledgment}\label{sec:acknowledgment}
This work was supported by the Institute of Information \& Communications Technology Planning \& Evaluation (IITP) grant funded by the Korea government (MSIT) (No. RS-2024-00457882, No. RS-2019-II191906, No. RS-2022-II220713, No. RS-2026-25531289), the National Research Foundation of Korea (NRF) grant funded by the Korea government (MSIT) (No. RS-2024-00453301, No. RS-2025-24873016, No. RS-2026-25494004), and the Basic Science Research Program through the National Research Foundation of Korea (NRF) funded by the Ministry of Education.

\newpage
{
\small
\bibliographystyle{ieeenat_fullname}
\bibliography{main}

@String(CVPR= {IEEE Conf. Comput. Vis. Pattern Recog.})

@String(CVPR  = {CVPR})

@inproceedings{
dosovitskiy2021an,
title={An Image is Worth 16x16 Words: Transformers for Image Recognition at Scale},
author={Alexey Dosovitskiy and Lucas Beyer and Alexander Kolesnikov and Dirk Weissenborn and Xiaohua Zhai and Thomas Unterthiner and Mostafa Dehghani and Matthias Minderer and Georg Heigold and Sylvain Gelly and Jakob Uszkoreit and Neil Houlsby},
booktitle={International Conference on Learning Representations},
year={2021},
}

@inproceedings{zhai2023siglip,
title={Sigmoid loss for language image pre-training},
author={Zhai, Xiaohua and Mustafa, Basil and Kolesnikov, Alexander and Beyer, Lucas},
booktitle={Proceedings of the IEEE/CVF international conference on computer vision},
year={2023}
}

@article{liu2021post,
title={Post-training quantization for vision transformer},
author={Liu, Zhenhua and Wang, Yunhe and Han, Kai and Zhang, Wei and Ma, Siwei and Gao, Wen},
journal={Advances in Neural Information Processing Systems},
year={2021}
}

@inproceedings{liu2023noisyquant,
title={{NoisyQuant}: Noisy bias-enhanced post-training activation quantization for vision transformers},
author={Liu, Yijiang and Yang, Huanrui and Dong, Zhen and Keutzer, Kurt and Du, Li and Zhang, Shanghang},
booktitle={Proceedings of the IEEE/CVF Conference on Computer Vision and Pattern Recognition},
year={2023}
}

@inproceedings{li2023repq,
title={{RepQ-ViT}: Scale reparameterization for post-training quantization of vision transformers},
author={Li, Zhikai and Xiao, Junrui and Yang, Lianwei and Gu, Qingyi},
booktitle={Proceedings of the IEEE/CVF International Conference on Computer Vision},
year={2023}
}

@inproceedings{yuan2022ptq4vit,
title={{PTQ4ViT}: Post-training quantization for vision transformers with twin uniform quantization},
author={Yuan, Zhihang and Xue, Chenhao and Chen, Yiqi and Wu, Qiang and Sun, Guangyu},
booktitle={European conference on computer vision},
year={2022}
}

@inproceedings{sun2024massive,
title={Massive Activations in Large Language Models},
author={Sun, Mingjie and Chen, Xinlei and Kolter, J Zico and Liu, Zhuang},
booktitle={Conference on Language Modeling},
year={2024}
}

@inproceedings{gu2025when,
title={When Attention Sink Emerges in Language Models: An Empirical View},
author={Xiangming Gu and Tianyu Pang and Chao Du and Qian Liu and Fengzhuo Zhang and Cunxiao Du and Ye Wang and Min Lin},
booktitle={International Conference on Learning Representations},
year={2025}
}

@inproceedings{son2024prefixing,
title={Prefixing Attention Sinks can Mitigate Activation Outliers for Large Language Model Quantization},
author={Son, Seungwoo and Park, Wonpyo and Han, Woohyun and Kim, Kyuyeun and Lee, Jaeho},
booktitle={Empirical Methods in Natural Language Processing},
year={2024}
}

@inproceedings{darcet2024vision,
title={Vision Transformers Need Registers},
author={Timoth{\'e}e Darcet and Maxime Oquab and Julien Mairal and Piotr Bojanowski},
booktitle={International Conference on Learning Representations},
year={2024}
}

@inproceedings{
jiang2025vision,
title={Vision Transformers Don't Need Trained Registers},
author={Nicholas Jiang and Amil Dravid and Alexei A Efros and Yossi Gandelsman},
booktitle={Advances in neural information processing systems},
year={2025},
}

@inproceedings{guo2024attention,
title={Attention Score is not All You Need for Token Importance Indicator in KV Cache Reduction: Value Also Matters},
author={Guo, Zhiyu and Kamigaito, Hidetaka and Watanabe, Taro},
booktitle={Empirical Methods in Natural Language Processing},
year={2024}
}

@article{oquab2024dinov,
title={{DINO}v2: Learning Robust Visual Features without Supervision},
author={Maxime Oquab and Timoth{\'e}e Darcet and Th{\'e}o Moutakanni and Huy V. Vo and Marc Szafraniec and Vasil Khalidov and Pierre Fernandez and Daniel HAZIZA and Francisco Massa and Alaaeldin El-Nouby and Mido Assran and Nicolas Ballas and Wojciech Galuba and Russell Howes and Po-Yao Huang and Shang-Wen Li and Ishan Misra and Michael Rabbat and Vasu Sharma and Gabriel Synnaeve and Hu Xu and Herve Jegou and Julien Mairal and Patrick Labatut and Armand Joulin and Piotr Bojanowski},
journal={Transactions on Machine Learning Research},
year={2024},
}

@inproceedings{cherti2023openclip,
title={Reproducible scaling laws for contrastive language-image learning},
author={Cherti, Mehdi and Beaumont, Romain and Wightman, Ross and Wortsman, Mitchell and Ilharco, Gabriel and Gordon, Cade and Schuhmann, Christoph and Schmidt, Ludwig and Jitsev, Jenia},
booktitle={Proceedings of the IEEE/CVF conference on computer vision and pattern recognition},
year={2023}
}

@article{tschannen2025siglip2,
title={{SigLIP} 2: Multilingual vision-language encoders with improved semantic understanding, localization, and dense features},
author={Tschannen, Michael and Gritsenko, Alexey and Wang, Xiao and Naeem, Muhammad Ferjad and Alabdulmohsin, Ibrahim and Parthasarathy, Nikhil and Evans, Talfan and Beyer, Lucas and Xia, Ye and Mustafa, Basil and others},
journal={arXiv preprint arXiv:2502.14786},
year={2025}
}

@inproceedings{radford2021learning,
title={Learning transferable visual models from natural language supervision},
author={Radford, Alec and Kim, Jong Wook and Hallacy, Chris and Ramesh, Aditya and Goh, Gabriel and Agarwal, Sandhini and Sastry, Girish and Askell, Amanda and Mishkin, Pamela and Clark, Jack and others},
booktitle={International conference on machine learning},
year={2021},
}

@inproceedings{deng2009imagenet,
title={{ImageNet}: A large-scale hierarchical image database},
author={Deng, Jia and Dong, Wei and Socher, Richard and Li, Li-Jia and Li, Kai and Fei-Fei, Li},
booktitle={IEEE conference on computer vision and pattern recognition},
year={2009},
}

@inproceedings{kang2025see,
title={See What You Are Told: Visual Attention Sink in Large Multimodal Models},
author={Seil Kang and Jinyeong Kim and Junhyeok Kim and Seong Jae Hwang},
booktitle={International Conference on Learning Representations},
year={2025},
}

@article{lu2025artifacts,
title={Artifacts and Attention Sinks: Structured Approximations for Efficient Vision Transformers},
author={Lu, Andrew and Liao, Wentinn and Wang, Liuhui and Yang, Huzheng and Shi, Jianbo},
journal={arXiv preprint arXiv:2507.16018},
year={2025}
}

@inproceedings{xiao2024efficient,
title={Efficient Streaming Language Models with Attention Sinks},
author={Guangxuan Xiao and Yuandong Tian and Beidi Chen and Song Han and Mike Lewis},
booktitle={International Conference on Learning Representations},
year={2024},
}

@article{chen2024prefixquant,
title={{PrefixQuant}: Eliminating outliers by prefixed tokens for large language models quantization},
author={Chen, Mengzhao and Liu, Yi and Wang, Jiahao and Bin, Yi and Shao, Wenqi and Luo, Ping},
journal={arXiv preprint arXiv:2410.05265},
year={2024}
}

@inproceedings{lin2014mscoco,
title={Microsoft coco: Common objects in context},
author={Lin, Tsung-Yi and Maire, Michael and Belongie, Serge and Hays, James and Perona, Pietro and Ramanan, Deva and Doll{\'a}r, Piotr and Zitnick, C Lawrence},
booktitle={European conference on computer vision},
year={2014},
}

@inproceedings{wu2025fima,
title={{FIMA-Q}: Post-Training Quantization for Vision Transformers by Fisher Information Matrix Approximation},
author={Wu, Zhuguanyu and Wang, Shihe and Zhang, Jiayi and Chen, Jiaxin and Wang, Yunhong},
booktitle={Proceedings of the Computer Vision and Pattern Recognition Conference},
year={2025}
}

@inproceedings{fei2004caltech101,
title={Learning generative visual models from few training examples: An incremental bayesian approach tested on 101 object categories},
author={Fei-Fei, Li and Fergus, Rob and Perona, Pietro},
booktitle={conference on computer vision and pattern recognition workshop},
year={2004},
}

@inproceedings{stanfordcar,
title={3d object representations for fine-grained categorization},
author={Krause, Jonathan and Stark, Michael and Deng, Jia and Fei-Fei, Li},
booktitle={Proceedings of the IEEE international conference on computer vision workshops},
year={2013}
}

@inproceedings{flowers,
title={Automated flower classification over a large number of classes},
author={Nilsback, Maria-Elena and Zisserman, Andrew},
booktitle={Indian conference on computer vision, graphics \& image processing},
year={2008},
}

@inproceedings{food,
title={Food-101--mining discriminative components with random forests},
author={Bossard, Lukas and Guillaumin, Matthieu and Van Gool, Luc},
booktitle={European conference on computer vision},
year={2014},
}

@article{dubey2024llama3,
title={The Llama 3 Herd of Models},
author={Dubey, Abhimanyu and Jauhri, Abhinav and Pandey, Abhishek and Kadian, Abhishek and Al-Dahle, Ahmad and Letman, Aiesha and Mathur, Akhil and Schelten, Alan and Yang, Aobo and Fan, Angela and others},
journal={arXiv preprint arXiv:2407.21783},
year={2024}
}

@inproceedings{choukroun2019low,
title={Low-bit quantization of neural networks for efficient inference},
author={Choukroun, Yoni and Kravchik, Eli and Yang, Fan and Kisilev, Pavel},
booktitle={IEEE/CVF International Conference on Computer Vision Workshop},
year={2019}
}

@inproceedings{dettmers2022gpt3,
title={{GPT3.int8}(): 8-bit matrix multiplication for transformers at scale},
author={Dettmers, Tim and Lewis, Mike and Belkada, Younes and Zettlemoyer, Luke},
booktitle={Advances in neural information processing systems},
year={2022}
}

@inproceedings{xiao2023smoothquant,
title={{SmoothQuant}: Accurate and efficient post-training quantization for large language models},
author={Xiao, Guangxuan and Lin, Ji and Seznec, Mickael and Wu, Hao and Demouth, Julien and Han, Song},
booktitle={International conference on Machine Learning},
year={2023},
}

@inproceedings{lin2024awq,
title={{AWQ}: Activation-aware weight quantization for on-device {LLM} compression and acceleration},
author={Lin, Ji and Tang, Jiaming and Tang, Haotian and Yang, Shang and Chen, Wei-Ming and Wang, Wei-Chen and Xiao, Guangxuan and Dang, Xingyu and Gan, Chuang and Han, Song},
booktitle={Proceedings of Machine Learning and Systems},
year={2024}
}

@article{yang2024mitigating,
title={Mitigating quantization errors due to activation spikes in {GLU}-based {LLMs}},
author={Yang, Jaewoo and Kim, Hayun and Kim, Younghoon},
journal={arXiv preprint 2405.14428},
year={2024}
}

@article{simeoni2025dinov3,
title={Dinov3},
author={Sim{\'e}oni, Oriane and Vo, Huy V and Seitzer, Maximilian and Baldassarre, Federico and Oquab, Maxime and Jose, Cijo and Khalidov, Vasil and Szafraniec, Marc and Yi, Seungeun and Ramamonjisoa, Micha{\"e}l and others},
journal={arXiv preprint arXiv:2508.10104},
year={2025}
}

@inproceedings{timkey2021all,
title={All bark and no bite: Rogue dimensions in transformer language models obscure representational quality},
author={Timkey, William and Van Schijndel, Marten},
booktitle={Empirical Methods in Natural Language Processing},
year={2021}
}

@inproceedings{kovaleva2021bert,
title={{BERT} busters: Outlier dimensions that disrupt transformers},
author={Kovaleva, Olga and Kulshreshtha, Saurabh and Rogers, Anna and Rumshisky, Anna},
booktitle={Findings of the ACL},
year={2021}
}

@inproceedings{bondarenko2021understanding,
title={Understanding and overcoming the challenges of efficient transformer quantization},
author={Bondarenko, Yelysei and Nagel, Markus and Blankevoort, Tijmen},
booktitle={Empirical Methods in Natural Language Processing},
year={2021}
}

@inproceedings{xiao2021noise,
title={Noise or Signal: The Role of Image Backgrounds in Object Recognition},
author={Kai Yuanqing Xiao and Logan Engstrom and Andrew Ilyas and Aleksander Madry},
booktitle={International Conference on Learning Representations},
year={2021},
}

@inproceedings{zhao2025viditq,
title={{ViDiT-Q}: Efficient and Accurate Quantization of Diffusion Transformers for Image and Video Generation},
author={Tianchen Zhao and Tongcheng Fang and Haofeng Huang and Rui Wan and Widyadewi Soedarmadji and Enshu Liu and Shiyao Li and Zinan Lin and Guohao Dai and Shengen Yan and Huazhong Yang and Xuefei Ning and Yu Wang},
booktitle={International Conference on Learning Representations},
year={2025}
}

@article{ashkboos2024quarot,
title={{QuaRot}: Outlier-free 4-bit inference in rotated llms},
author={Ashkboos, Saleh and Mohtashami, Amirkeivan and Croci, Maximilian L and Li, Bo and Cameron, Pashmina and Jaggi, Martin and Alistarh, Dan and Hoefler, Torsten and Hensman, James},
journal={Advances in Neural Information Processing Systems},
year={2024}
}

@inproceedings{liu2024spinquant,
title={{SpinQuant}: {LLM} Quantization with Learned Rotations},
author={Zechun Liu and Changsheng Zhao and Igor Fedorov and Bilge Soran and Dhruv Choudhary and Raghuraman Krishnamoorthi and Vikas Chandra and Yuandong Tian and Tijmen Blankevoort},
booktitle={International Conference on Learning Representations},
year={2025}
}

@article{zhong2025erq,
title={Towards Accurate Post-Training Quantization of Vision Transformers via Error Reduction},
author={Zhong, Yunshan and Huang, You and Hu, Jiawei and Zhang, Yuxin and Ji, Rongrong},
journal={IEEE Trans. Pattern Anal. Mach. Intell.},
year={2025}
}

@inproceedings{
bolya2025perception,
title={Perception Encoder: The best visual embeddings are not at the output of the network},
author={Daniel Bolya and Po-Yao Huang and Peize Sun and Jang Hyun Cho and Andrea Madotto and Chen Wei and Tengyu Ma and Jiale Zhi and Jathushan Rajasegaran and Hanoona Abdul Rasheed and Junke Wang and Marco Monteiro and Hu Xu and Shiyu Dong and Nikhila Ravi and Shang-Wen Li and Piotr Dollar and Christoph Feichtenhofer},
booktitle={The Thirty-ninth Annual Conference on Neural Information Processing Systems},
year={2025},
}

@inproceedings{zhou2017scene,
title={Scene Parsing through ADE20K Dataset},
author={Zhou, Bolei and Zhao, Hang and Puig, Xavier and Fidler, Sanja and Barriuso, Adela and Torralba, Antonio},
booktitle={Proceedings of the IEEE Conference on Computer Vision and Pattern Recognition},
year={2017}
}

@inproceedings{parkhi12oxfordpet,
author={Omkar M. Parkhi and Andrea Vedaldi and Andrew Zisserman and C. V. Jawahar},
title={Cats and Dogs},
booktitle={IEEE Conference on Computer Vision and Pattern Recognition},
year={2012},
}

@inproceedings{cimpoi14dtd,
Author={M. Cimpoi and S. Maji and I. Kokkinos and S. Mohamed and A. Vedaldi},
Title={Describing Textures in the Wild},
Booktitle={Proceedings of the {IEEE} Conf. on Computer Vision and Pattern Recognition ({CVPR})},
Year={2014}
}

@article{
faghri2025mobileclip,
title={Mobile{CLIP}2: Improving Multi-Modal Reinforced Training},
author={Fartash Faghri and Pavan Kumar Anasosalu Vasu and Cem Koc and Vaishaal Shankar and Alexander T Toshev and Oncel Tuzel and Hadi Pouransari},
journal={Transactions on Machine Learning Research},
year={2025},
}

@article{liu2023visual,
  title={Visual instruction tuning},
  author={Liu, Haotian and Li, Chunyuan and Wu, Qingyang and Lee, Yong Jae},
  journal={Advances in neural information processing systems},
  year={2023}
}

@article{beyer2024paligemma,
  title={Paligemma: A versatile 3b vlm for transfer},
  author={Beyer, Lucas and Steiner, Andreas and Pinto, Andr{\'e} Susano and Kolesnikov, Alexander and Wang, Xiao and Salz, Daniel and Neumann, Maxim and Alabdulmohsin, Ibrahim and Tschannen, Michael and Bugliarello, Emanuele and others},
  journal={arXiv preprint arXiv:2407.07726},
  year={2024}
}

@inproceedings{vasu2025fastvlm,
  title={Fastvlm: Efficient vision encoding for vision language models},
  author={Vasu, Pavan Kumar Anasosalu and Faghri, Fartash and Li, Chun-Liang and Koc, Cem and True, Nate and Antony, Albert and Santhanam, Gokula and Gabriel, James and Grasch, Peter and Tuzel, Oncel and others},
  booktitle={Proceedings of the Computer Vision and Pattern Recognition Conference},
  year={2025}
}

@inproceedings{li2024llama,
  title={Llama-vid: An image is worth 2 tokens in large language models},
  author={Li, Yanwei and Wang, Chengyao and Jia, Jiaya},
  booktitle={European Conference on Computer Vision},
  year={2024}
}

@article{tai2024mptq,
  title={{MPTQ-ViT}: Mixed-precision post-training quantization for vision transformer},
  author={Tai, Yu-Shan and others},
  journal={arXiv preprint arXiv:2401.14895},
  year={2024}
}

@article{bai2025qwen3,
  title={Qwen3-vl technical report},
  author={Bai, Shuai and Cai, Yuxuan and Chen, Ruizhe and Chen, Keqin and Chen, Xionghui and Cheng, Zesen and Deng, Lianghao and Ding, Wei and Gao, Chang and Ge, Chunjiang and others},
  journal={arXiv preprint arXiv:2511.21631},
  year={2025}
}

@article{hudson2018gqa,
    title={{GQA}: A New Dataset for Real-World Visual Reasoning 
    and Compositional Question Answering},
    author={Hudson, Drew A and Manning, Christopher D},
    journal={Conference on Computer Vision and Pattern Recognition (CVPR)},
    year={2019}
}

@InProceedings{goyal2017vqav2,
    author = {Yash Goyal and Tejas Khot and Douglas Summers{-}Stay and Dhruv Batra and Devi Parikh},
    title = {Making the {V} in {VQA} Matter: Elevating the Role of Image Understanding in {V}isual {Q}uestion {A}nswering},
    booktitle = {Conference on Computer Vision and Pattern Recognition (CVPR)},
    year = {2017},
}

@inproceedings{fu2025videomme,
  title={{Video-MME}: The first-ever comprehensive evaluation benchmark of multi-modal llms in video analysis},
  author={Fu, Chaoyou and Dai, Yuhan and Luo, Yongdong and Li, Lei and Ren, Shuhuai and Zhang, Renrui and Wang, Zihan and Zhou, Chenyu and Shen, Yunhang and Zhang, Mengdan and others},
  booktitle={CVPR},
  year={2025}
}

@inproceedings{zhou2025mlvu,
  title={{MLVU}: Benchmarking multi-task long video understanding},
  author={Zhou, Junjie and Shu, Yan and Zhao, Bo and Wu, Boya and Liang, Zhengyang and Xiao, Shitao and Qin, Minghao and Yang, Xi and Xiong, Yongping and Zhang, Bo and others},
  booktitle={Proceedings of the IEEE/CVF Conference on Computer Vision and Pattern Recognition},
  year={2025}
}

@misc{trtll,
  author       = {{NVIDIA Corporation}},
  title        = {Deploying Hugging Face Llava1.5-7b Model in Triton},
  howpublished = {\url{https://docs.nvidia.com/deeplearning/triton-inference-server/user-guide/docs/tutorials/Popular_Models_Guide/Llava1.5/llava_trtllm_guide.html}},
}

@misc{trt,
  author       = {{NVIDIA Corporation}},
  title        = {NVIDIA TensorRT},
  howpublished = {\url{https://developer.nvidia.com/tensorrt}},
}
}

\appendix
\clearpage
\onecolumn  
\appendix
\begingroup
\small                   
\linespread{1.15}\selectfont     
\setlength{\parindent}{0pt}

\begin{center}
{\LARGE\bfseries Appendix\par}
\end{center}

\vspace{1.2em}

{\large\bfseries Contents\par}
\vspace{0.9em}
\hrule
\vspace{0.9em}

\appTocSec{app:qsen}
\appTocSec{app:test_time_register}
\appTocSec{app:app_rtrv}
\appTocSec{app:app_vlm}

\appTocSec{app:app_baselines}
\appTocSubManual{E.1}{app:e_models}{Models}
\appTocSubManual{E.2}{app:e_baselines}{Baselines}
\appTocSubManual{E.3}{app:e_hyperparameters}{Hyperparameters}
\appTocSubManual{E.4}{app:e_preprocessing_cost}{Preprocessing}
\appTocSubManual{E.5}{app:e_hardware}{Hardware}

\appTocSec{app:cossim}
\appTocSec{app:app_vizreg}
\appTocSec{app:simpler}
\appTocSec{app:segmentation}
\appTocSec{app:outlier_impact}
\appTocSec{app:hp_analysis}
\appTocSec{app:recon_loss}
\appTocSec{app:fp16token}
\appTocSec{app:freq}
\appTocSec{app:app_woq}
\appTocSec{app:hadamard}
\appTocSec{app:SmoothQuant}

\endgroup

\vspace{1em}

\clearpage

\section{Additional Results on Quantization Sensitivity}\label{app:qsen}
\begin{figure}[h]
\begin{center}
\begin{minipage}{1.0\linewidth}
\centering
\includegraphics[width=0.85\textwidth]{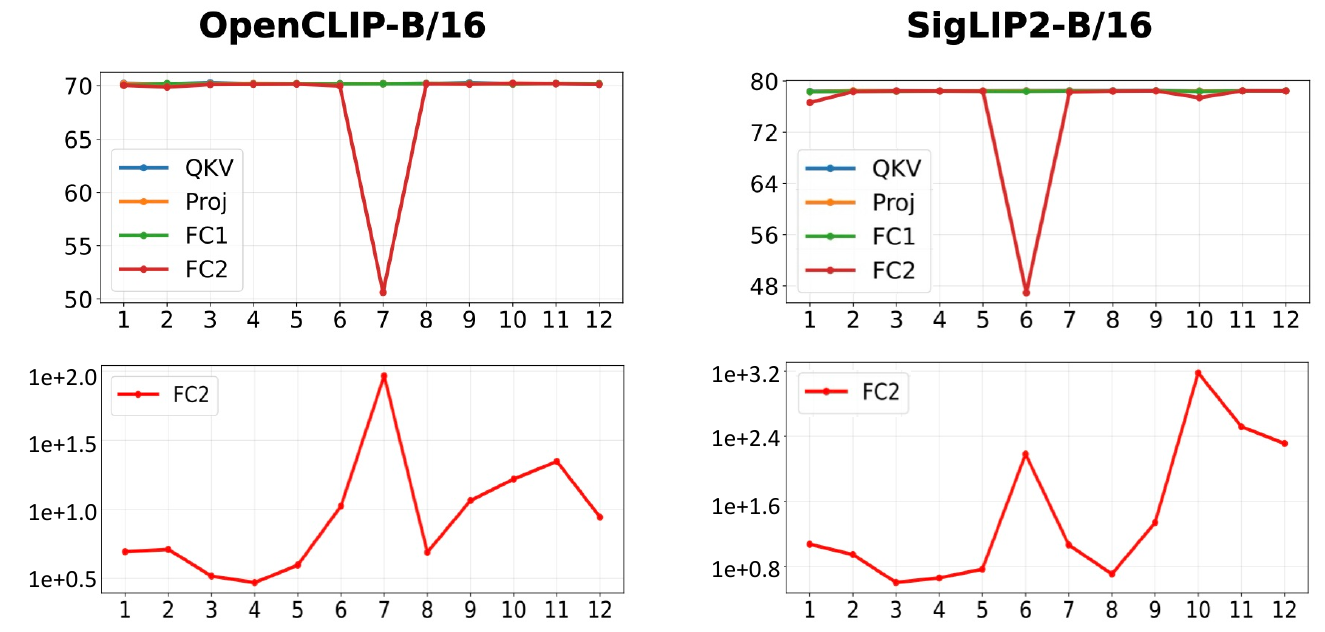}
\end{minipage}
\caption{\textbf{
(Top) Layerwise quantization sensitivity (\%).} Zero-shot ImageNet-1k accuracy when we quantize only one layer to W8A8. \textbf{(Bottom) Layerwise max token norms.} The largest $\ell_\infty$-norm of all tokens in an image on a logarithmic scale, averaged over the ImageNet-1k validation set.}
\vspace{-4.5em}
\label{fig:quantization_sensitivity_app}
\end{center}
\end{figure}

In this section, we provide additional plots for other vision encoders, analogous to those in \Cref{fig:quantization_sensitivity}. In \cref{fig:quantization_sensitivity_app}, we plot layerwise quantization sensitivity (top row) and maximum token norm (bottom row) for OpenCLIP and SigLIP2. The trends are consistent with our analysis in \cref{ssec:qsen}: increases in maximum token norm coincide with decreases in quantized accuracy. However, in the case of SigLIP2, the absolute scale of the maximum norm is significantly larger than in the other architectures we considered. Consequently, applying RegCache yields a clearer benefit, as shown in \cref{tab:naive_best_config_with_diff}. 
Given SigLIP2's distinct behavior compared to other vision encoders, it would be intriguing to investigate further; we leave this for future work.

\newpage
\section{Comparison with Test-Time Registers}\label{app:test_time_register}
Our experimental results demonstrate that RegCache consistently outperforms the test-time register \cite{jiang2025vision} variant in quantization settings. As discussed in \cref{sec:related_work}, Jiang et al.~\cite{jiang2025vision} propose register tokens as a concurrent approach to mitigating activation outliers. In an extended experiment reported in Appendix 2 (Ver.~5 \cite{jiang2025vision}), they further inject test-time registers as a form of KV cache while zeroing out outlier neurons, showing that high-norm residual-stream outliers can be suppressed. To assess the practical utility of this mechanism in quantized models, we evaluate it under identical conditions and compare it directly with RegCache.

The results show that RegCache remains consistently superior, suggesting that identifying the optimal curation tailored to a quantization-based setup is more effective than relying on test-time register based outlier suppression mechanism.
\begin{table}[h]
\centering
\caption{\textbf{Comparison with test-time register.} Accuracy of zero-shot image classification on ImageNet-1k under W8A8 quantization.}
\label{tab:test_time_register}
\setlength{\tabcolsep}{8pt}
\resizebox{0.40\linewidth}{!}{
\begin{tabular}{lcc}
\toprule
\textbf{Method} & \textbf{CLIP-B/16} & \textbf{SigLIP-B/16} \\
\midrule
FP32                & 68.32 & 76.05 \\
\midrule
Na\"{i}ve                & 34.01 & 69.71 \\
Test-time register        & 44.30 & 73.81 \\
RegCache                  & \textbf{59.71} & \textbf{74.38} \\
\bottomrule
\end{tabular}
}
\end{table}

\newpage

\section{Additional Results on Image Retrieval}\label{app:app_rtrv}
We present supplementary tables for additional vision encoders, in a format consistent with \cref{tab:retrieval_main}. For other models, \cref{tab:retrieval_app} presents zero-shot image–text retrieval results on the MS-COCO dataset for OpenCLIP and SigLIP2. In both models, RegCache consistently achieves the highest retrieval accuracy as well. 
\begin{table*}[ht!]
\small
\centering
\caption{\textbf{Zero-shot image--text retrieval performance of OpenCLIP and SigLIP2 on MS-COCO (R@1).} The best results are marked in \textbf{bold}. Best/Average $\Delta$ denote the maximum/average accuracy gaps between each baseline and its RegCache counterpart, excluding the Na\"ive cases.}
\renewcommand{\arraystretch}{1.15}
\setlength{\tabcolsep}{4pt}
\resizebox{0.8\linewidth}{!}{
\begin{tabular}{lcccccccccccc}
\toprule
& \multicolumn{6}{c}{\textbf{OpenCLIP-B/16}} & \multicolumn{6}{c}{\textbf{SigLIP2-B/16}} \\
\cmidrule(lr){2-7}\cmidrule(lr){8-13}
& \multicolumn{3}{c}{\textbf{Image $\rightarrow$ Text}} & \multicolumn{3}{c}{\textbf{Text $\rightarrow$ Image}}
& \multicolumn{3}{c}{\textbf{Image $\rightarrow$ Text}} & \multicolumn{3}{c}{\textbf{Text $\rightarrow$ Image}} \\
\cmidrule(lr){2-4}\cmidrule(lr){5-7}\cmidrule(lr){8-10}\cmidrule(lr){11-13}
& W4A4 & W6A6 & W8A8
& W4A4 & W6A6 & W8A8
& W4A4 & W6A6 & W8A8
& W4A4 & W6A6 & W8A8 \\
\midrule

\multicolumn{1}{l}{FP32}
& 61.02 & 61.02 & 61.02
& 41.38 & 41.38 & 41.38
& 71.60 & 71.60 & 71.60
& 52.33 & 52.33 & 52.33 \\
\midrule

Na\"ive
& 0.02 & 0.16 & 37.32
& 0.03 & 0.24 & 26.30
& 0.02 & 0.02 & 14.26
& 0.02 & 0.16 & 13.86 \\
\rowcolor{gray!10}
w/ RegCache
& 0.00 & 0.32 & 57.60
& 0.04 & 0.77 & 38.45
& 0.02 & 0.08 & 64.02
& 0.03 & 0.22 & 46.80 \\
\midrule

PTQ4ViT
& 0.04 & 50.60 & 59.60
& 0.06 & 34.56 & 40.66
& 0.04 & 29.88 & 69.74
& 0.19 & 27.33 & 51.62 \\
\rowcolor{gray!10}
w/ RegCache
& 1.00 & \textbf{60.00} & \textbf{60.00}
& 3.00 & \textbf{40.96} & \textbf{40.96}
& 0.70 & 53.90 & 70.12
& 2.64 & 39.96 & 51.68 \\
\midrule

RepQ-ViT
& 0.12 & 17.60 & 57.62
& 0.21 & 8.90 & 38.88
& 0.16 & 57.00 & 69.20
& 0.28 & 39.20 & 50.15 \\
\rowcolor{gray!10}
w/ RegCache
& 1.56 & 32.96 & 59.44
& 1.38 & 14.90 & 39.82
& 3.20 & 58.78 & 69.76
& 3.49 & 41.26 & 50.15 \\
\midrule

NoisyQuant
& 0.46 & 43.86 & 53.84
& 1.20 & 27.05 & 34.50
& 0.07 & 28.74 & 62.04
& 0.65 & 25.28 & 46.32 \\
\rowcolor{gray!10}
w/ RegCache
& 2.18 & 51.58 & 56.06
& 2.81 & 31.73 & 35.53
& 0.32 & 50.66 & 70.24
& 1.79 & 39.25 & 51.41 \\
\midrule

FIMA-Q
& 55.82 & 58.84 & 59.26
& 38.02 & 40.39 & 40.21
& 67.70 & 67.68 & 67.70
& 47.22 & 47.21 & 47.22 \\
\rowcolor{gray!10}
w/ RegCache
& \textbf{57.34} & 59.88 & 59.86
& \textbf{40.02} & 40.30 & 40.58
& \textbf{71.54} & \textbf{71.54} & \textbf{71.52}
& \textbf{52.49} & \textbf{52.49} & \textbf{52.50} \\
\midrule

ERQ
& 9.70 & 32.56 & 58.74
& 4.28 & 14.74 & 39.83
& 8.28 & 64.80 & 69.76
& 5.78 & 45.44 & 49.97 \\
\rowcolor{gray!10}
w/ RegCache
& 13.02 & 33.64 & 59.56
& 5.19 & 15.38 & 40.04
& 12.30 & 65.56 & 70.00
& 9.18 & 46.28 & 50.20 \\
\midrule

Best $\Delta$
& \incr{+3.32} & \incr{+15.36} & \incr{+2.22}
& \incr{+2.94} & \incr{+6.40} & \incr{+1.03}
& \incr{+4.02} & \incr{+24.02} & \incr{+8.20}
& \incr{+5.27} & \incr{+13.97} & \incr{+5.28} \\
Average $\Delta$
& \incr{+1.79} & \incr{+6.92} & \incr{+1.17}
& \incr{+1.73} & \incr{+3.53} & \incr{+0.57}
& \incr{+2.36} & \incr{+10.47} & \incr{+2.64}
& \incr{+3.09} & \incr{+6.96} & \incr{+2.13} \\
\bottomrule
\end{tabular}
}
\label{tab:retrieval_app}
\end{table*}

\newpage

% \section{Additional results of VLM}\label{sec:app_vlm}
% We further evaluate RegCache on the widely adopted LLaVA family \cite{liu2024llava, liu2024llavanext} using the GQA and VQAv2 benchmarks, as reported in \cref{tab:vlm_performance_app}. In \cref{tab:vlm_performance}, we report the results under a practical deployment-oriented setup, i.e., 4-bit quantization with PTQ baseline. We note that, RepQ-ViT is a practical method among our baselines, since it appears good performance improvement and affordable calibration time and memory for more large size of vision encoders. Consistent with our observations on Qwen3, \cite{bai2025qwen3} RegCache also provides consistent gains in this setting, suggesting that its effectiveness is not limited to a single backbone.
\section{Additional Results on VLMs}\label{app:app_vlm}
We further evaluate RegCache on the widely adopted LLaVA family on the GQA and VQAv2 benchmarks, as summarized in \cref{tab:vlm_performance_app}. In \cref{tab:vlm_performance}, we consider a practical deployment setting with 4-bit quantization and RepQ-ViT as the PTQ baseline. We note that RepQ-ViT is particularly effective in practice, offering both favorable accuracy gains and low calibration overhead, especially for larger vision encoders in VLMs. Consistent with our observations on Qwen3-VL \cite{bai2025qwen3}, RegCache delivers stable gains in this setting, confirming that its efficacy extends beyond a specific architecture.

\begin{table*}[h]
    \centering
    \setlength{\tabcolsep}{6pt}
    \renewcommand{\arraystretch}{1.15}
    \caption{\textbf{VLM performance on LLaVA architecture-based models under 4-bit quantization.}}
    \label{tab:vlm_performance_app}
    \begin{subtable}{\linewidth}
        \centering
        \caption{\textbf{LLaVA-1.5 models.}}
        \label{tab:vlm_performance_app_llava15}
        \resizebox{0.6\linewidth}{!}{%
        \begin{tabular}{lcccc}
        \toprule
        \multirow{2}{*}{\textbf{Method}} &
        \multicolumn{2}{c}{\textbf{LLaVA-1.5-7B}} &
        \multicolumn{2}{c}{\textbf{LLaVA-1.5-13B}} \\
        \cmidrule(lr){2-3}\cmidrule(lr){4-5}
        &
        \textbf{GQA} & \textbf{VQAv2} &
        \textbf{GQA} & \textbf{VQAv2} \\
        \midrule
        FP32
            & 60.78 & 76.64
            & 62.57 & 78.29 \\
        \midrule
        Na\"ive
            & 33.84 & 35.62
            & 34.70 & 35.89 \\
        RepQ-ViT
            & 41.40 & 45.87
            & 43.27 & 47.85 \\
        w/ RegCache
            & 45.10 \incr{(+3.70)} & 50.11 \incr{(+4.24)}
            & 46.21 \incr{(+2.94)} & 52.15 \incr{(+4.30)} \\
        \bottomrule
        \end{tabular}%
        }
    \end{subtable}

    \vspace{1em}

    \begin{subtable}{\linewidth}
        \centering
        \caption{\textbf{LLaVA-NeXT models.}}
        \label{tab:vlm_performance_app_llava16}
        \resizebox{0.6\linewidth}{!}{%
        \begin{tabular}{lcccc}
        \toprule
        \multirow{2}{*}{\textbf{Method}} &
        \multicolumn{2}{c}{\textbf{LLaVA-1.6-vicuna-13B}} &
        \multicolumn{2}{c}{\textbf{LLaVA-1.6-mistral-7B}} \\
        \cmidrule(lr){2-3}\cmidrule(lr){4-5}
        &
        \textbf{GQA} & \textbf{VQAv2} &
        \textbf{GQA} & \textbf{VQAv2} \\
        \midrule
        FP32
            & 63.08 & 79.33
            & 56.59 & 76.15 \\
        \midrule
        Na\"ive
            & 35.72 & 36.70
            & 31.87 & 35.21 \\
        RepQ-ViT
            & 48.93 & 56.70
            & 44.63 & 53.45 \\
        w/ RegCache
            & 51.40 \incr{(+2.47)} & 58.83 \incr{(+2.13)}
            & 46.21 \incr{(+1.58)} & 55.64 \incr{(+2.19)} \\
        \bottomrule
        \end{tabular}%
        }
    \end{subtable}
\end{table*}

\newpage

\section{Experimental Settings and Baselines}\label{app:app_baselines}
In this section, we provide additional details on the vision encoder models, quantization baselines, and the hardware setup used in our experiments.

% \paragraph{\textbf{Models.}}
\subsection{Models}\label{app:e_models}
The vision encoders we choose cover a broad range of training objectives (image--text contrastive learning vs.\ self-supervised learning) and architectural choices for global feature aggregation, providing a diverse testbed for evaluating RegCache. CLIP and SigLIP are trained on image--text pairs with contrastive objectives, whereas DINOv2 is trained on image-only datasets. Regarding input tokens, CLIP and DINOv2 utilize a class token to extract global features, while SigLIP and SigLIP2 use patch-wise pooling to generate a token that captures global information. 

% \paragraph{\textbf{Baselines.}}
\subsection{Baselines}\label{app:e_baselines}
We consider the following baselines ranging from widely used to recent:
\begin{itemize}[leftmargin=*,topsep=0pt,parsep=1pt]
\item \textbf{PTQ4ViT} \cite{yuan2022ptq4vit} proposes a twin uniform quantizer to handle the unbalanced activation distributions found in ViTs, particularly after non-linearities such as Softmax and GELU.
\item \textbf{RepQ-ViT} \cite{li2023repq} addresses quantization bottlenecks by applying specialized preprocessing to sensitive layers, such as channel-wise quantization after LayerNorm and log2 quantization after Softmax.
\item \textbf{NoisyQuant} \cite{liu2023noisyquant} introduces a quantizer-agnostic strategy that adds fixed uniform bias to activations, thereby reducing the quantization error of heavy-tailed distributions.
\item \textbf{FIMA-Q} \cite{wu2025fima} suggests optimizing the rounding function and scaling factor using effective Hessian-guided quantization loss.
\item \textbf{ERQ} \cite{zhong2025erq} decouples activation and weight error reduction via Ridge-regression-based weight correction. It also integrates several effective PTQ techniques, including incorporating channel-wise scaling factors into the LayerNorm parameters to handle activation outliers, and applying group-wise uniform quantization with separate quantizers for weight outliers, along with rounding function optimization.
\end{itemize}
All baselines use per-tensor dynamic quantization with 8-/6-/4-bit integer precision. Following the original papers, we use 1,024 calibration samples for NoisyQuant and 32 calibration samples for RepQ-ViT, PTQ4ViT, and ERQ. For FIMA-Q, we use 1,024 samples for optimization and 128 samples for calibration.

\clearpage 

\subsection{Hyperparameters}\label{app:e_hyperparameters}
For each vision encoder and PTQ baseline, we search over the hyperparameters of RegCache to obtain the best-performing configuration. The default search ranges used in our experiments are summarized in Table~\ref{tab:hyperparams}.

\begin{table}[H]
\centering
\caption{\textbf{Default search ranges.} We search for the register extraction layer starting from the quantization-sensitive layer, and sequentially sweep the number of prefixed and deleted tokens.}
\label{tab:hyperparams}
\begingroup
\footnotesize
\renewcommand{\arraystretch}{0.8}
\setlength{\tabcolsep}{12pt}
\begin{tabular}{lc}
\toprule
\textbf{Hyperparameter} & \textbf{Search range} \\
\midrule
Register extraction layer & 4 \\
Max Prefix tokens & 15 \\
Max Deleted tokens & 10 \\
\bottomrule
\end{tabular}
\endgroup
\end{table}

\subsection{Preprocessing}\label{app:e_preprocessing_cost}
We further note that the preprocessing is performed for each model and quantization setup (i.e., PTQ baseline and bit-width), while the resulting configuration can be transferred to other tasks and datasets.

\subsection{Hardware}\label{app:e_hardware}
Most standalone vision encoder experiments are conducted on NVIDIA RTX 4090 GPUs, whereas VLM evaluations are performed on NVIDIA RTX A6000 GPUs. Due to its large memory requirements, the FIMA-Q baseline is evaluated on both NVIDIA RTX A6000 and NVIDIA A100 80GB GPUs.

\newpage

\section{High Cosine Similarity of Outlier Tokens}\label{app:cossim}

In this section, we provide a simple explanation for why outlier tokens maintain high cosine similarity across images, as shown in \cref{tab:cosine_similarity}.

As a stylized example, consider two distinct outlier tokens, each modeled as a ``spiked'' vector in which a single element has high magnitude (i.e., $C$ in \cref{lem1}). \cref{lem1} says that the high cosine similarity over the outlier tokens can largely be
attributed to two factors\footnote{There are some cases where each component has a small magnitude, yet the vectors are still similar to one another. In high dimensions, however, this rarely happens; for example, random vectors become almost orthogonal as the dimension tends to infinity. In vision encoders, the number of channels per token is typically large (e.g., 768 channels in CLIP and SigLIP families), which results in low similarity among non-outlier tokens (e.g., ``Normal tokens'' in \cref{tab:cosine_similarity}).}: \textbf{(i) the presence of large-magnitude entries} and
\textbf{(ii) the alignment of the corresponding entry indices (i.e., same or at least similar channels)}.

\begin{lemma}\label{lem1}
    Let $\mathbf{x}, \mathbf{y} \in \mathbb{R}^d$ and fix an index $i \in \{1,\dots,d\}$.
    Moreover, let $\mathbf{1}_i \in \mathbb{R}^d$ denote the one-hot vector whose $i$th entry is $1$
    and all other entries are $0$.
    For $C \in \mathbb{R}^+$, define $
        \mathbf{x}' = \mathbf{x} + C \mathbf{1}_i,~
        \mathbf{y}' = \mathbf{y} + C \mathbf{1}_i. $
    Then the vectors asymptotically converge to each other, i.e.,
    \begin{align}
        \lim_{C\rightarrow\infty}
        \frac{\langle \mathbf{x}', \mathbf{y}'\rangle}
             {\|\mathbf{x}'\|_2 \,\|\mathbf{y}'\|_2}
        = 1. \label{eq:same}
    \end{align}
    Moreover, for  $
        \mathbf{y}'' = \mathbf{y} + C \mathbf{1}_j,$ where $j\neq i$, then the vectors become asymptotically orthogonal to each other, i.e.,
        \begin{align}
        \lim_{C\rightarrow\infty}
        \frac{\langle \mathbf{x}', \mathbf{y}''\rangle}
             {\|\mathbf{x}'\|_2 \,\|\mathbf{y}''\|_2}
        = 0. \label{eq:diff}
        \end{align}
\end{lemma}
\begin{proof}
For the same-index case (\cref{eq:same}), we can write this explicitly as
\begin{align}
    \frac{\langle \mathbf{x}', \mathbf{y}'\rangle}
         {\|\mathbf{x}'\|_2 \,\|\mathbf{y}'\|_2}
    =
    \frac{\langle \mathbf{x}, \mathbf{y}\rangle 
          + C(x_i + y_i) + C^2}
         {\sqrt{\|\mathbf{x}\|_2^2 + 2C x_i + C^2}\,
          \sqrt{\|\mathbf{y}\|_2^2 + 2C y_i + C^2}},
\end{align}
where $x_i$ (resp. $y_i$) denotes the $i$th entry of $\mathbf x$ (resp. $\mathbf y$). Dividing numerator and denominator by $C^2$, we get
\begin{align}
    \frac{
        1 + \dfrac{x_i + y_i}{C} + \dfrac{\langle \mathbf{x}, \mathbf{y}\rangle}{C^2}
    }{
        \sqrt{1 + \dfrac{2x_i}{C} + \dfrac{\|\mathbf{x}\|_2^2}{C^2}}
        \sqrt{1 + \dfrac{2y_i}{C} + \dfrac{\|\mathbf{y}\|_2^2}{C^2}}
    }.
\end{align}
As $C \to \infty$, all terms of order $1/C$ and $1/C^2$ vanish, and we get what we want. The case of the different-index (\cref{eq:diff}) can be handled similarly.
\end{proof}

Indeed, as shown in \cref{fig:placeholder} (right), we find that across different images, outlier tokens tend to have similar magnitudes in certain coordinates (i.e., along specific channels shared across different tokens).

\newpage

\begin{figure*} [h]
    \centering
    \includegraphics[width=1\textwidth]{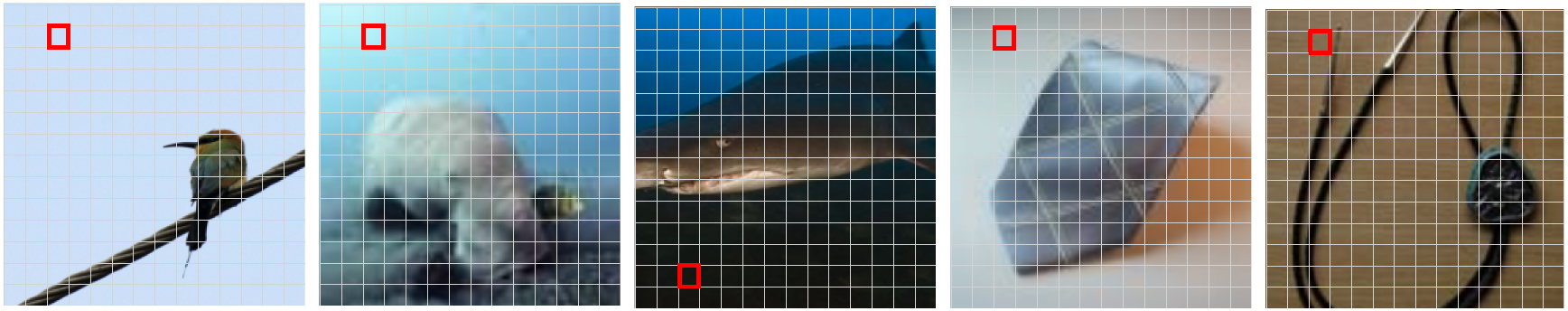}
    \caption{Visualization of curated prefix tokens from ImageNet-1k}
    \label{fig:reg_tokens}
\end{figure*}

\section{Visualization of Curated Prefix Tokens}\label{app:app_vizreg}
\noindent In \cref{fig:reg_tokens}, we visualize the top 5 prefix tokens selected by the method described in \cref{ssec:curating}, ranked by their effectiveness under quantization as measured by W8A8 zero-shot classification accuracy on ImageNet-1k. The results are consistent with prior findings~\cite{darcet2024vision, jiang2025vision}, revealing that the register tokens are located in background regions. We find that the selected register tokens commonly correspond to low-frequency regions surrounded by semantically uninformative patches. 

\newpage

\section{Comparison with Simpler Baselines}\label{app:simpler}

We compare RegCache against two simpler heuristics under W8A8 quantization on
CLIP-B/16: \textbf{(i) Random prefixing}, which prefixes a randomly selected
sink patch token instead of the averaged curated registers; and
\textbf{(ii) Outlier clipping}, which directly clips the outliers at the
quantization-sensitive layer to the mean norm of the remaining tokens, narrowing
the dynamic range without adding or removing any tokens.

As shown in Tab.~\ref{tab:simpler}, both improve over na\"ive quantization but
fall clearly short of RegCache. Random prefixing recovers only part of the
accuracy, indicating that a single sink token lacks the generality of averaged
registers, while outlier clipping helps even less, as clipping to a fixed norm threshold distorts the functional role of high-magnitude activations along with the outliers. These results confirm that the gains of RegCache stem from the curation and
averaging of sink-like register tokens, rather than from generic outlier suppression.

\begin{table}[h]
\centering
\caption{\textbf{Simpler outlier-mitigation strategies on CLIP-B/16.}
Zero-shot ImageNet-1k accuracy (\%) under W8A8 quantization.}
\label{tab:simpler}
\begin{tabular}{lc}
\toprule
\textbf{Method} & \textbf{Accuracy} \\
\midrule
FP32 & 68.32 \\
\midrule
Na\"ive & 34.01 \\
w/ Random prefixing & 54.18~(\incr{+20.17}) \\
w/ Outlier clipping & 47.29~(\incr{+13.28}) \\
w/ RegCache & \textbf{59.71}~(\incr{+25.70}) \\
\bottomrule
\end{tabular}
\end{table}
\newpage

\section{Segmentation Results}\label{app:segmentation}
To analyze the impact of token deletion on dense prediction, we conduct semantic segmentation experiments on ADE20K\protect\footnotemark ~and report the mIoU results in \cref{tab:sink_analysis}. We find that the sink token shows a noticeably lower mIoU score, consistent with prior observations that sink tokens can degrade segmentation performance \cite{jiang2025vision,bolya2025perception}. To remove the sink token while maintaining spatial structure, we replace its position with an interpolated token obtained by averaging the surrounding patch embeddings, and report the resulting performance in \cref{tab:segment_w_reg}. Our method effectively suppresses the outliers induced by sink-token deletion, thereby improving post-training quantization performance on dense prediction tasks.
\footnotetext{We use ADE20K-\textit{MIT Scene Parsing Challenge 2016}, which is the subset of the ADE20K benchmark \cite{zhou2017scene}, available at \url{http://sceneparsing.csail.mit.edu/}.}
\begin{table}[h]
\centering
\caption{\textbf{Analysis of sink tokens and RegCache for segmentation.} Results on ADE20K using W8A8 quantized SigLIP-B/16.}\label{tab:seg_and_retrieval}
% ---------------- Left: Segmentation ----------------
\begin{subtable}[h]{0.35\columnwidth}
\centering
\caption{\textbf{mIoU by token}}
\label{tab:sink_analysis}
\setlength{\tabcolsep}{8pt}
\resizebox{0.55\columnwidth}{!}{
\begin{tabular}{lc}
\toprule
\textbf{Setting} & \textbf{mIoU} \\
\midrule
Sink          & 16.67 \\
Random        & 12.02 \\
Non-sink      & 30.15 \\
\bottomrule
\end{tabular}
}
\end{subtable}
% ---------------- Right: Retrieval / table in the image ----------------
\begin{subtable}[h]{0.60\columnwidth}
\centering
\caption{\textbf{Increased mIoU with RegCache}}
\label{tab:segment_w_reg}
\setlength{\tabcolsep}{8pt}
\resizebox{0.45\columnwidth}{!}{
\begin{tabular}{lc}
\toprule
\textbf{Method} & \textbf{mIoU} \\
\midrule
FP32 & 32.85 \\
\midrule
Na\"ive & 30.30 \\
RegCache w/ interp. & \textbf{32.46} \\
\bottomrule
\end{tabular}
}
\end{subtable}
\end{table}

\newpage

% \vspace{-0.5em}
\section{Impact of Activation Outlier on Quantization}\label{app:outlier_impact}
To assess the impact of outlier on quantization, we control the quantization range at the token level. In \cref{tab:outlier_impact_appn}, ``w/ outlier'' applies standard per-tensor quantization. In contrast, the ``w/o outlier'' setting excludes outlier tokens from the per-tensor range and quantizes them separately via per-token quantization. As a result, quantizing the outlier tokens separately shows a substantial performance improvement, which demonstrates that mitigating outliers is a critical problem in quantization. 
\begin{table}[h]
\centering
\caption{Comparison of quantization performance w/ and w/o outlier.}
\vspace{-0.7em}
\label{tab:outlier_impact_appn}
\setlength{\tabcolsep}{8pt}
\resizebox{0.6\columnwidth}{!}{
\begin{tabular}{lccc}
\toprule
\textbf{Model} & \textbf{Full precision} & \textbf{w/ outlier} & \textbf{w/o outlier} \\
\midrule
CLIP-B/16      & 68.32 & 34.01 & 55.83 (\incr{+21.82}) \\
OpenCLIP-B/16  & 70.22 & 46.12 & 65.43 (\incr{+19.31}) \\
SigLIP-B/16    & 76.05 & 69.71 & 74.54 (\incr{+4.83})  \\
SigLIP2-B/16   & 78.47 & 26.04 & 74.54 (\incr{+48.50}) \\
DINOv2-B/14    & 83.26 & 19.20 & 76.58 (\incr{+57.38}) \\
\bottomrule
\end{tabular}
}
\end{table}

\newpage

\section{Hyperparameter-Sensitivity Analysis}\label{app:hp_analysis}
\begin{figure}[t]
\centering

% ---------- First row: centered 2 plots ----------
\makebox[\textwidth][c]{%
\begin{subfigure}[t]{0.31\textwidth}
    \centering
    \includegraphics[width=\linewidth]{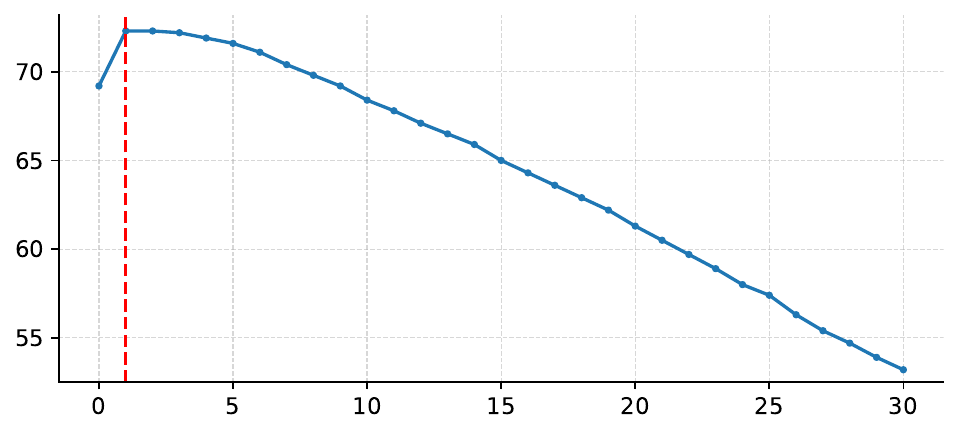}
    \caption{}
    \label{fig:hp_num_prefix}
\end{subfigure}
\hspace{0.08\textwidth}
\begin{subfigure}[t]{0.31\textwidth}
    \centering
    \includegraphics[width=\linewidth]{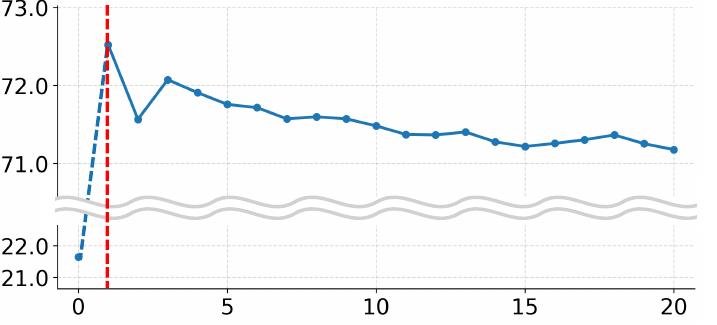}
    \caption{}
    \label{fig:hp_num_deleted}
\end{subfigure}%
}

\vspace{0.6em}

% ---------- Second row: 3 plots ----------
\begin{subfigure}[t]{0.31\textwidth}
    \centering
    \includegraphics[width=\linewidth]{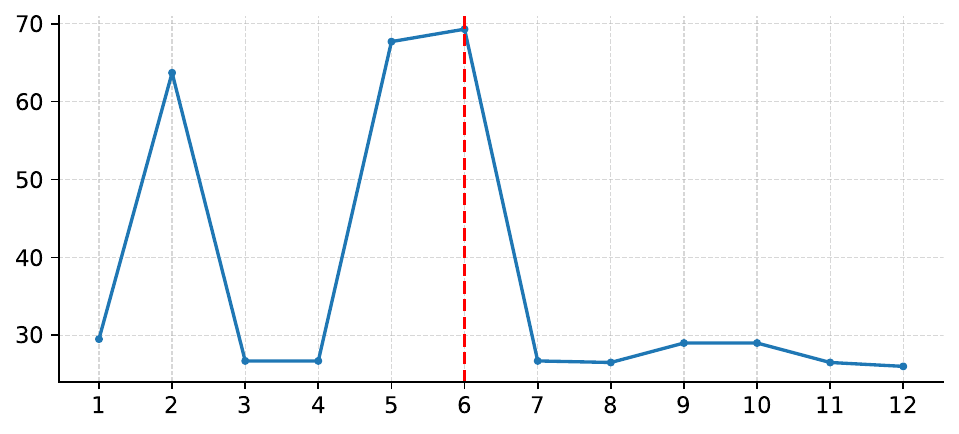}
    \caption{}
    \label{fig:hp_selected_layer}
\end{subfigure}
\hfill
\begin{subfigure}[t]{0.31\textwidth}
    \centering
    \includegraphics[width=\linewidth]{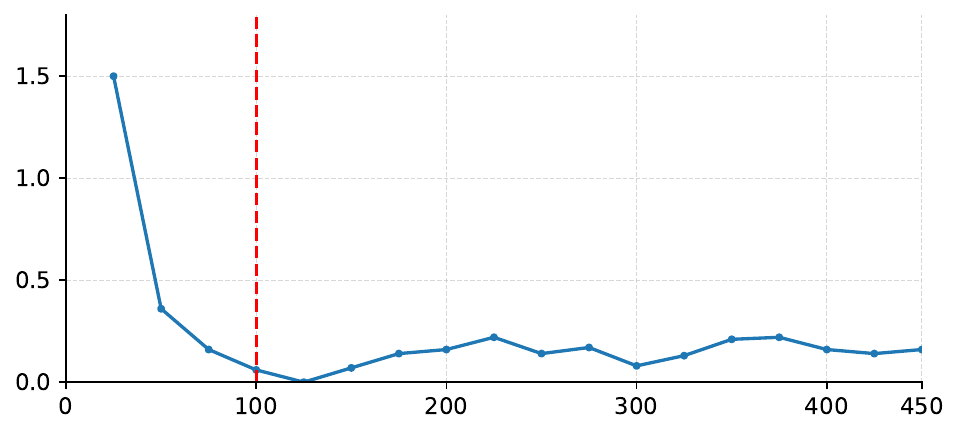}
    \caption{}
    \label{fig:hp_num_candidates}
\end{subfigure}
\hfill
\begin{subfigure}[t]{0.31\textwidth}
    \centering
    \includegraphics[width=\linewidth]{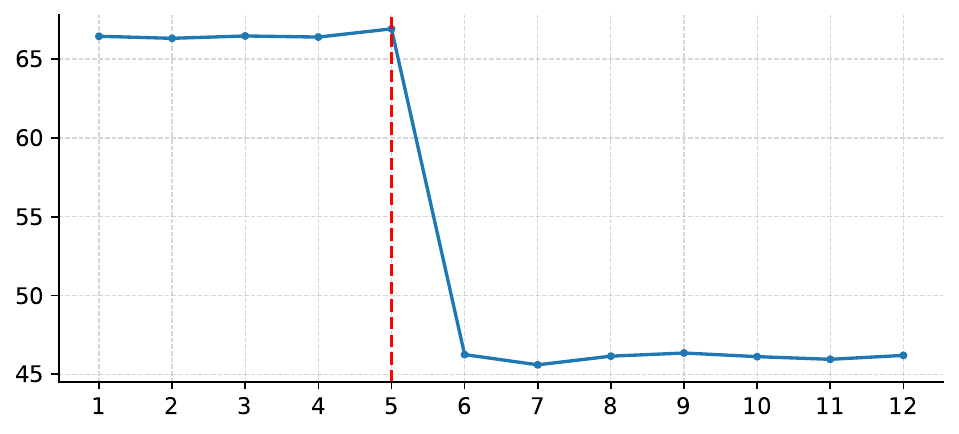}
    \caption{}
    \label{fig:hp_insertion_layer}
\end{subfigure}

\caption[Parameter-sensitivity analysis]{\textbf{Parameter-sensitivity analysis.} We evaluate our method on ImageNet-1k zero-shot classification under W8A8 quantization on SigLIP2, sweeping one parameter at a time: (a) the number of prefix tokens, (b) the number of deleted tokens, (c) the selection of the critical layer, (d) the candidate-pool size, i.e., the number of candidate tokens averaged, and (e) the prefix-insertion start layer. Plots (a)--(c) and (e) report accuracy, while (d) reports the percentage \textit{drop} from the peak performance. Red dashed lines indicate the settings used in the main experiments.}

\label{fig:parameter_analysis}
\end{figure}
 
\cref{fig:parameter_analysis} presents the hyperparameter analysis described above. \cref{fig:hp_num_prefix} and \cref{fig:hp_num_deleted} show that RegCache is robust to the number of prefix tokens and the number of deleted tokens, respectively. In both cases, each curve exhibits a clear sweet spot: performance first improves as our proposed components take effect, then declines as excessive prefix tokens or token deletion introduces noise or causes information loss. \cref{fig:hp_selected_layer} shows that the quantization-sensitive layer must be chosen carefully through search, since selecting other layers degrades performance substantially. \cref{fig:hp_num_candidates} indicates that our choice of 100 candidate tokens is the smallest value at which performance plateaus near its maximum. Finally, \cref{fig:hp_insertion_layer} shows how accuracy varies with the layer at which prefix insertion begins. Accuracy drops sharply when insertion starts at layer 6, immediately after the quantization-sensitive layer, indicating that prefix tokens are effective only when inserted before the sensitive layer. Based on this observation, we begin prefix insertion at an intermediate layer preceding the sensitive layer, reducing computational cost while preserving effectiveness.

\newpage

\section{RegCache Selected Based on Reconstruction Loss}\label{app:recon_loss}
To further demonstrate the extensibility of our approach, we additionally evaluate a quantization reconstruction loss–based search as an alternative selection strategy. Specifically, we search for the number of prefix and deleted tokens that minimize the MSE between output features produced by FP32 processing and those produced under the quantized setting. This criterion depends only on internal activations and does not require access to target-domain labels or downstream metrics, enabling a lightweight and broadly applicable procedure. 

The results in \cref{tab:recon_loss} show that our method remains effective under this label-free selection rule, supporting its practical deployment in scenarios where validation data is unavailable or restricted. Moreover, even without ImageNet-1k-based tuning, our method improves accuracy on average across quantization settings.

\begin{table*}[h]
\small
\centering
\caption{\textbf{Results of reconstruction loss--based search.} Zero-shot classification accuracy on ImageNet-1k for various vision encoders. The best results are marked in \textbf{bold}. Best/Average $\Delta$ denote the gaps between the best/average performance of each baseline with and without RegCache, excluding the Na\"ive cases. Here, RegCache {\textcolor{gray}{\scriptsize rc}} denotes the results with RegCache selected based on reconstruction loss.}
\renewcommand{\arraystretch}{1.0}
\setlength{\tabcolsep}{5pt}
\resizebox{0.5\linewidth}{!}{
\begin{tabular}{l ccc ccc}
\toprule
& \multicolumn{3}{c}{\textbf{CLIP-B/16}} 
& \multicolumn{3}{c}{\textbf{SigLIP-B/16}} \\
\cmidrule(lr){2-4} \cmidrule(lr){5-7}
\textbf{Method}
& \textbf{W4A4} & \textbf{W6A6} & \textbf{W8A8} 
& \textbf{W4A4} & \textbf{W6A6} & \textbf{W8A8} \\
\midrule
FP32 
& \multicolumn{3}{c}{68.32}
& \multicolumn{3}{c}{76.05} \\
\midrule
Na\"ive 
& 0.09 & 0.17 & 34.01
& 0.13 & 0.77 & 69.71 \\
\rowcolor{gray!10}
w/ RegCache {\textcolor{gray}{\scriptsize rc}}
& 0.09 & 0.27 & 59.47
& 0.13 & 23.29 & 74.31 \\
\midrule
PTQ4ViT 
& 0.37 & 51.60 & 67.69
& 0.19 & 68.68 & 75.57 \\
\rowcolor{gray!10}
w/ RegCache {\textcolor{gray}{\scriptsize rc}}
& 0.32 & 58.33 & 67.77
& 2.19 & 72.27 & 75.93 \\
\midrule
RepQ-ViT 
& 1.83 & 53.25 & 67.39
& 21.36 & 73.32 & 75.23 \\
\rowcolor{gray!10}
w/ RegCache {\textcolor{gray}{\scriptsize rc}}
& 15.09 & 66.81 & 68.00
& 30.91 & 74.59 & 75.99 \\
\midrule
NoisyQuant 
& 0.34 & 46.19 & 63.20
& 1.78 & 71.10 & 75.50 \\
\rowcolor{gray!10}
w/ RegCache {\textcolor{gray}{\scriptsize rc}}
& 1.29 & 57.19 & 65.80
& 8.61 & 72.18 & 75.69 \\
\midrule
FIMA-Q 
& 50.41 & 66.51 & 67.63
& 76.05 & 76.06 & 76.06 \\
\rowcolor{gray!10}
w/ RegCache {\textcolor{gray}{\scriptsize rc}}
& \textbf{62.27} & \textbf{66.90} & 67.78
& \textbf{76.12} & \textbf{76.12} & \textbf{76.12} \\
\midrule
ERQ 
& 1.56 & 39.64 & 67.99
& 57.76 & 74.44 & 75.95 \\
\rowcolor{gray!10}
w/ RegCache {\textcolor{gray}{\scriptsize rc}}
& 47.11 & 65.90 & \textbf{68.03}
& 60.06 & 75.24 & 76.04 \\
\midrule
Best $\Delta$
& \incr{+45.55} & \incr{+26.26} & \incr{+2.18}
& \incr{+9.55} & \incr{+3.59} & \incr{+0.76} \\
Average $\Delta$
& \incr{+14.32} & \incr{+11.59} & \incr{+0.61}
& \incr{+4.15} & \incr{+1.36} & \incr{+0.29} \\
\bottomrule
\end{tabular}
}
\label{tab:recon_loss}
\end{table*}
\clearpage
\newpage

\section{Using FP16 Instead of Token Deletion}\label{app:fp16token}
As a flexible variant of our method, we replace token deletion with FP16 token computation and report the results in \cref{tab:fp16token}. This option yields a modest accuracy improvement, but introduces a bottleneck in computing the FP16 tokens to propagate, thereby slowing down the overall process.

\begin{table}[h]
\centering
\caption{\textbf{Performance of FP16 mixed precision instead of token deleting.} Accuracy of zero-shot image classification on ImageNet-1k under W8A8 quantization.}
\label{tab:fp16token}
\setlength{\tabcolsep}{8pt}
\resizebox{0.40\linewidth}{!}{
\begin{tabular}{lcc}
\toprule
\textbf{Method} & \textbf{SigLIP} & \textbf{SigLIP2} \\
\midrule
FP32                        & 76.05    & 78.47    \\
\midrule
Na\"ive                     & 69.71 & 26.04 \\
w/ RegCache                 & 74.38 & 72.35 \\
w/ RegCache (FP16 MP)       & 74.51 & 72.82 \\
\bottomrule
\end{tabular}
}
\end{table}

\newpage

\begin{figure*}[h]
\begin{center}
\centering
\includegraphics[width=\textwidth]{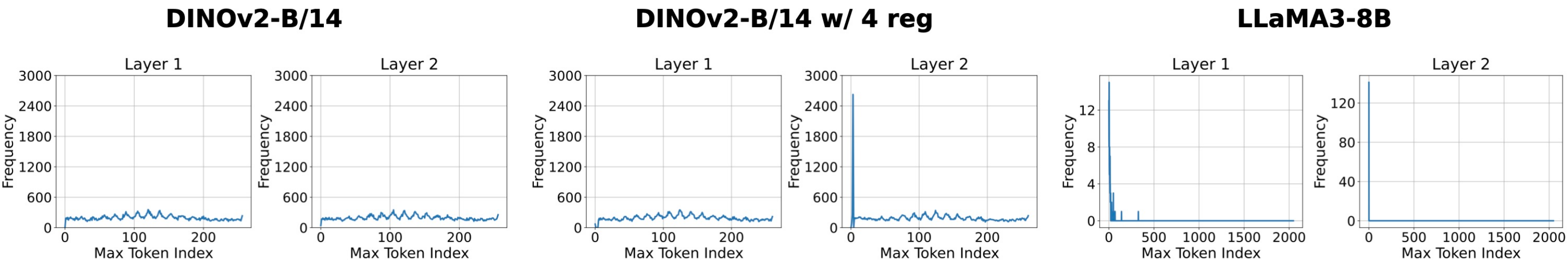}
\end{center}
\caption{\textbf{The frequency of top-1 max tokens in the input tensor of FC2 layers in different models}. We evaluated DINOv2 on ImageNet-1k and LLaMA3-8B on WikiText-2 dataset.}
\label{fig:vit_closedset}
\end{figure*}

% \vspace{-3em}
\section{Outliers in Vision Encoders vs. LLMs: A Tokenization Perspective}\label{app:freq}
As shown in \cref{fig:quantization_sensitivity}, various vision encoders exhibit outliers in the intermediate layers. Rewinding recent studies about outlier tendency in LLMs \cite{sun2024massive, gu2025when}, it is natural to ask: why do outliers consistently emerge in the intermediate layers of vision encoders, rather than in the early layers as in LLMs? In this section, we investigate this phenomenon by reconsidering the difference of \textit{tokenization strategies} between vision encoders and LLMs. 

Roughly speaking, LLMs map input sequences to tokens drawn from a discrete, fixed vocabulary. In contrast, vision encoders and other ViT-based models process inputs by mapping them to continuous token embeddings using a (convolutional) neural network. We hypothesize that differences in the emergence of sink tokens can be attributed to their fundamentally distinct tokenization process.

To test this hypothesis, we compare the outlier behavior of DINOv2, pretrained both with and without learned register tokens, and LLaMA3-8B \cite{dubey2024llama3}. In this setup, the register tokens act as ``fixed outlier sinks'' \cite{darcet2024vision}, effectively forming a closed-set vocabulary for outlier attraction, analogous to the tokenization setup in LLMs. As shown in \cref{fig:vit_closedset}, when ViTs are equipped with four learned register tokens, they begin to exhibit outliers in early layers (i.e., 2nd layer), mirroring the behavior observed in LLMs. This supports our hypothesis that continuous tokenization in ViTs plays a crucial role in the emergence of outliers in the intermediate layers. 

\newpage

\section{Weight-only Quantization}\label{app:app_woq}
\noindent We demonstrate the effectiveness of RegCache when combined with weight-centric methods, i.e., weight-only quantization, which are commonly used in LLMs to reduce memory usage and deployment cost. Specifically, we adopt AWQ~\cite{lin2024awq}, a widely used weight-only quantization method, as the baseline, using a group size of 128 and bitwidths of 8, 6, 4, and 3. Across all configurations, RegCache consistently improves performance over vanilla AWQ, demonstrating its complementary benefits even in memory-constrained quantization settings. However, as noted in \cref{sec:introduction}, unlike autoregressive LLMs, vision encoders are typically compute-bound, making weight-only quantization less effective.
\begin{table*}[h]
\centering
\footnotesize
\renewcommand{\arraystretch}{1.15}
\caption{Zero-shot image classification accuracy (\%) under weight-only quantization (AWQ).}
\label{tab:awq_regcache_rowwise}
\setlength{\tabcolsep}{8pt}
\resizebox{0.80\linewidth}{!}{
\begin{tabular}{l c l cccc}
\toprule
\multirow{2}{*}{\textbf{Model}} & \multirow{2}{*}{\textbf{FP32}} & \multirow{2}{*}{\textbf{Method}} & \multicolumn{4}{c}{\textbf{Weight-only (AWQ) Bits}} \\
\cmidrule(lr){4-7}
 &  &  & \textbf{W3A16} & \textbf{W4A16} & \textbf{W6A16} & \textbf{W8A16} \\
\midrule
\multirow{2}{*}{\textbf{CLIP-B/16}} & \multirow{2}{*}{68.32} & AWQ
  & 62.08 & 66.73 & 67.80 & 68.05 \\
 &  & \cellcolor{gray!10} + RegCache
  & \cellcolor{gray!10} 63.08 \inc{+1.00} & \cellcolor{gray!10} 67.06 \inc{+0.33} & \cellcolor{gray!10} 67.94 \inc{+0.14} & \cellcolor{gray!10} 68.14 \inc{+0.09} \\
\bottomrule
\end{tabular}
}
\end{table*} 

\newpage

\section{With Hadamard Rotation Quantization}\label{app:hadamard}
Motivated by recent works in the LLM literature on utilizing Hadamard rotations for outlier mitigation~\cite{ashkboos2024quarot, liu2024spinquant}, we have also explored whether RegCache can be combined with such methods. In particular, we follow the adaptations of ViDiT-Q \cite{zhao2025viditq} to apply Hadamard rotations to all linear layers in the transformer blocks.\footnote{There is one exception: we do not use the Walsh–Hadamard matrix for the value and output projection matrices, as in the ViDiT-Q implementation. Also, we have empirically observed that na\"{i}vely applying QuaRot to vision encoders tends to severely damage the model performance.} As reported in \cref{tab:hadamard}, RegCache improves accuracy in most settings when used as a plug-in method, indicating that suppressing extreme outliers in the rotated domain can further enhance quantization performance.
\begin{table}[h]
\centering
\caption{Zero-shot image classification accuracy (\%) on top of QuaRot \cite{ashkboos2024quarot}}\label{tab:hadamard}
\vspace{-0.7em}
\setlength{\tabcolsep}{8pt}
\resizebox{0.45\linewidth}{!}{
\begin{tabular}{@{}lcccc@{}}
\toprule
\textit{}            & \multicolumn{2}{c}{\textbf{CLIP}}    & \multicolumn{2}{c}{\textbf{OpenCLIP}}      \\ 
\cmidrule(l){2-5} 
\textbf{Method}      & \textbf{W6A6} & \textbf{W8A8} & \textbf{W6A6} & \textbf{W8A8} \\ 
\midrule
FP32                   &  \multicolumn{2}{c}{68.32}       & \multicolumn{2}{c}{70.22}    \\
\midrule
Na\"{i}ve                & 0.17          & 34.01         & 0.47          & 46.12 \\
QuaRot             & 55.86  & 66.38 & 59.49 & \textbf{68.50} \\
QuaRot + RegCache  & \textbf{56.05} & \textbf{66.84} & \textbf{59.76} & \textbf{68.50} \\
\bottomrule
\end{tabular}}
\end{table}

\newpage

\section{Combining with LLM Outlier-mitigation Methods}\label{app:SmoothQuant}
We evaluate whether SmoothQuant (SQ)---a widely used outlier-mitigation PTQ technique for LLMs---transfers to ViT-based vision encoders. As shown in \cref{tab:smoothquant}, SQ alone yields only marginal improvements over the naïve quantization baseline, consistent with prior observations on ViTs~\cite{tai2024mptq}. While RegCache provides clear additional gains when applied on top of SQ, we further note that ViT-specific outlier mitigation methods such as RepQ-ViT (RepQ) may be a more effective direction for vision encoders.
\begin{table}[h]
\centering
\caption{\textbf{W8A8 quantization on vision encoders.} Zero-shot ImageNet-1k accuracy (\%) under W8A8 quantization.}
\label{tab:smoothquant}
\setlength{\tabcolsep}{8pt}
\resizebox{0.70\linewidth}{!}{
\begin{tabular}{l c c cc cc}
\toprule
& & & \multicolumn{2}{c}{\textbf{LLM PTQ method}} & \multicolumn{2}{c}{\textbf{ViT PTQ method}} \\
\cmidrule(lr){4-5} \cmidrule(lr){6-7}
 & \textbf{FP32} & \textbf{Na\"{\i}ve} & \textbf{SQ} & \textbf{SQ + ours} & \textbf{RepQ} & \textbf{RepQ + ours} \\
\midrule
\textbf{CLIP}      & 68.32 & 34.01 & 35.54 & 60.30~(\incr{+24.76}) & 67.39 & 68.10~(\incr{+0.71}) \\
\textbf{OpenCLIP}  & 70.22 & 46.12 & 46.50 & 67.07~(\incr{+20.57}) & 68.70 & 70.06~(\incr{+1.36}) \\
\textbf{SigLIP}    & 76.05 & 69.71 & 70.55 & 74.65~(\incr{+4.10})  & 75.23 & 75.94~(\incr{+0.71}) \\
\textbf{SigLIP2}   & 78.47 & 26.04 & 28.20 & 72.45~(\incr{+44.25}) & 76.43 & 77.13~(\incr{+0.70}) \\
\bottomrule
\end{tabular}
}
\end{table}

\end{document}